%% file: conf25-german-commons-frame.tex
\documentclass[sigconf,natbib=false,screen,authorversion,nonacm,review=false,timestamp=false]{acmart} % For preprint

\usepackage{conf25-german-commons}
\begin{document}
\input{conf25-german-commons-pre}
\input{conf25-german-commons-part1}
\input{conf25-german-commons-part2}
\input{conf25-german-commons-part3}

\input{conf25-german-commons-part4}
\input{conf25-german-commons-part5}
\input{conf25-german-commons-part6}
\input{conf25-german-commons-sum}

\balance
\printbibliography

\newpage
\appendix
\onecolumn
\section{Data Filtering}
\subsection{Quality Filtering Parameter Values}\label{app:data-filtering-parameters}
\input{table-filter-parameters}
\subsection{Stopwords}\label{app:data-filtering-stopwords}
\input{table-stopwords}
\subsection{PII Generic Replacement Values}\label{app:data-filtering-pii}
\input{table-pii-replacements}
\subsection{Processing Step Statistics}
\input{table-filtering-steps}\label{app:data-filtering-steps}

{
    \thispagestyle{empty}
    \titleformat*{\section}{\LARGE\bfseries}
    \titleformat{\subsection}[display]
        {\bfseries\normalsize}
        {\thesubsection}
        {2ex}
        {\vskip .1ex\titlerule\filright}
        [\titlerule\vskip .75ex]
    \titlespacing{\subsection}
        {0pt}{1ex plus .2ex minus .3ex}{0pt}
    \titleformat{\paragraph}[runin]
        {\bfseries}
        {\theparagraph}{1ex}{}
    \titlespacing{\paragraph}
        {0pt}{.5ex plus .1ex minus .2ex}{2pt}
    \sffamily
    \footnotesize
    \setlength\columnsep{6pt}
    \begin{multicols}{3}
    \input{datasheet}
    \end{multicols}
}
\end{document}

%% file: conf25-german-commons-pre.tex
\title[The German Commons -- 154 Billion Tokens of Openly Licensed Text for German Language Models]{The German Commons -- 154 Billion Tokens of \\ Openly Licensed Text for German Language Models}

%\title[German Commons -- Large-Scale, Diverse, Permissively Licensed Data for Open German Language Models]{German Commons -- Large-Scale, Diverse, Permissively \\ Licensed Data for Open German Language Models}
%\title{German Commons -- Large-Scale, Permissively Licensed, High-Quality Data for Open German Language Models}
%\title{\emojilizard Gecco -- A Family of German Language Models trained on Permissively Licensed Data}
%\title{\emojilizard Gecco -- A Family of German CC-licensed Open Models}
% \title{\emojilizard Gecco -- German CC-compliant Open Language Models} 
% Gecco -> *Ge*rman, *CC*-Licensed, *O*pen Models

\settopmatter{authorsperrow=4}

%% Author ordering not final

\author[L.~Gienapp]{Lukas Gienapp}
\affiliation{
  \institution{University of Kassel, hessian.AI, and ScaDS.AI}
  \city{Kassel}
  \country{Germany}
}
%\email{lukas.gienapp@uni-kassel.de}
\orcid{0000-0001-5707-3751}

\author[C.~Schröder]{Christopher Schröder}
\affiliation{
  \institution{InfAI and ScaDS.AI}
  \city{Leipzig} 
  \country{Germany}
}
\orcid{0000-0002-7081-8495}

\author[S.~Schweter]{Stefan Schweter}
\affiliation{
  \institution{Independent Researcher}
  \city{Holzkirchen} 
  \country{Germany}
}
\orcid{0000-0002-7190-2090}

\author[C.~Akiki]{Christopher Akiki}
\affiliation{
  \institution{Leipzig University and ScaDS.AI}
  \city{Leipzig} 
  \country{Germany}
}
\orcid{0000-0000-0000-0000}

\author[F.~Schlatt]{Ferdinand Schlatt}
\affiliation{
  \institution{Friedrich-Schiller-Universität Jena}
  \city{Jena} 
  \country{Germany}
}
\orcid{0000-0000-0000-0000}

\author[A.~Zimmermann]{Arden Zimmermann}
\affiliation{
  \institution{German National Library}
  \city{Leipzig} 
  \country{Germany} 
}
\orcid{0000-0000-0000-0000}

\author[P.~Gen\^et]{Philippe Gen\^et}
\affiliation{
  \institution{German National Library}
  \city{Frankfurt} 
  \country{Germany}
}
\orcid{0000-0000-0000-0000}

\author[M.~Potthast]{Martin Potthast}
\affiliation{
  \institution{University of Kassel, hessian.AI, and ScaDS.AI}
  \city{Kassel}
  \country{Germany}
}
%\email{martin.potthast@uni-kassel.de}
\orcid{0000-0003-2451-0665}

\renewcommand{\shortauthors}{Gienapp et al.}

\begin{abstract}
Large language model development relies on large-scale training corpora, yet most contain data of unclear licensing status, limiting the development of truly open models. This problem is exacerbated for non-English languages, where openly licensed text remains critically scarce. We introduce the \emph{German Commons}, the largest collection of openly licensed German text to date. It compiles data from \gcNumSources{}~sources across seven domains, encompassing legal, scientific, cultural, political, news, economic, and web text. Through systematic sourcing from established data providers with verifiable licensing, it yields \gcNumTokensInBillions{}~billion tokens of high-quality text for language model training. Our processing pipeline implements comprehensive quality filtering, deduplication, and text formatting fixes, ensuring consistent quality across heterogeneous text sources. All domain subsets feature licenses of at least CC-BY-SA~4.0 or equivalent, ensuring legal compliance for model training and redistribution. The \emph{German Commons} therefore addresses the critical gap in openly licensed German pretraining data, and enables the development of truly open German language models. We also release code for corpus construction and data filtering tailored to German language text, rendering the \emph{German Commons} fully reproducible and extensible.\\[.5\baselineskip]
{\small
\hbox to 3em {\bfseries Data:}  \url{https://huggingface.co/datasets/coral-nlp/german-commons} \\
\hbox to 3em {\bfseries Code:}  \url{https://github.com/coral-nlp/llmdata}                     \\
}
\end{abstract}

\keywords{}

%%%%%%%%%%%%%%%%%%%%%%%%%%%%%%%%%%%%%%%%%%%%%%%%%%%%%%%%%%%%%%%%%%%%%%%%%%%

% German Commons statistics
\newcommand{\gcNumSources}{41}
\newcommand{\gcNumDocsInMillions}{35.78}
\newcommand{\gcNumTokensInBillions}{154.56}

% Itemization.
\newcommand{\Ni}{(1)~}
\newcommand{\Nii}{(2)~}
\newcommand{\Niii}{(3)~}
\newcommand{\Niv}{(4)~}
\newcommand{\Nv}{(5)~}
\newcommand{\Nvi}{(6)~}

%%%%%%%%%%%%%%%%%%%%%%%%%%%%%%%%%%%%%%%%%%%%%%%%%%%%%%%%%%%%%%%%%%%%%%%%%%%
%%% The following automatically sets the ACM meta variables to reserve 
%%% sufficient space, and depending on whether the anonymous switch is set
%%% in documentclass.
%%%%%%%%%%%%%%%%%%%%%%%%%%%%%%%%%%%%%%%%%%%%%%%%%%%%%%%%%%%%%%%%%%%%%%%%%%%
\copyrightyear{2025}
\acmYear{2025}
\setcopyright{rightsretained}
\settopmatter{printacmref=false}

%%%%%%%%%%%%%%%%%%%%%%%%%%%%%%%%%%%%%%%%%%%%%%%%%%%%%%%%%%%%%%%%%%%%%%%%%%%

% The following includes the CC license icon appropriate for your paper.
% Download the image from www.scomminc.com/pp/acmsig/4ACM-CC-by-88x31.eps
% and place within your figs or figures folder

\makeatletter
\gdef\@copyrightpermission{
 \begin{minipage}{0.3\columnwidth}
  \href{https://creativecommons.org/licenses/by/4.0/}{\includegraphics[width=0.90\textwidth]{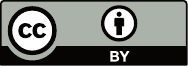}}
 \end{minipage}\hfill
 \begin{minipage}{0.7\columnwidth}
  \href{https://creativecommons.org/licenses/by/4.0/}{This work is licensed under a Creative Commons Attribution International 4.0 License.}
 \end{minipage}
}
\makeatother

\maketitle

%% file: conf25-german-commons-part1.tex
\section{Introduction}

%%% Starting point:  open models are great!
%Open language models are increasingly rivaling commercial systems in terms of their effectiveness and/or efficiency on key benchmarks, with expanding coverage of languages and tasks. For non-English, low-resource languages in particular, community initiatives are leading the development of tailored models.

Open language models are increasingly rivaling commercial systems in terms of their effectiveness and/or efficiency on key benchmarks, with expanding coverage of languages and tasks. Yet, the degree of openness of open models is often lacking~\cite{longpre:2023}. While the weights of open models are being released under open licenses, the licensing of the training data of many models remains unclear. However, as most pretraining datasets used are derived from large-scale web crawls, this creates legal and ethical barriers to the development of fully open language models, despite the web's long-established value in science~\cite{kilgarriff:2003} and industry~\cite{brin:1998}. This is because 
\Ni 
the provenance of web content is hard to establish and
\Nii
obtaining consent from original authors or copyright holders is infeasible at web scale; 
\Niii
re-publishing web data without such consent as part of a training dataset infringes upon copyright and privacy; in particular since
\Niv
the data may contain personally identifiable information~(PII), and
\Nv
generative models, even if published openly, may reproduce sensitive or copyrighted text.

These shortcomings limit the usefulness of many open-weight models. Practitioners can only trust, but not verify, that, for example, the web crawls used for training have not been contaminated by benchmark test data or that no copyrighted material is included. To minimize these risks and make open language models usable for both scientific and commercial purposes without reservations, it is important to limit their training data to verifiably open texts. However, this is challenging for non-English languages, as the available open alternatives to web data need to be carefully compiled.
 
\begin{table}[t]
\centering
\renewcommand{\arraystretch}{1.2}
\caption{Overview of number of documents, number of tokens and median sequence length per source domain comprised in the German Commons corpus.}
\label{tab:overview}
\input{table-overview}

\end{table}

%%% Problem: Open-weight-only models.
%%% NOTES.
%%% - Previous attempt at the second paragraph. 
%%% - It again is just a defense of (true) openness, not leading to our contributions. 
%%% - Potentially integrate in a later section.
%However, many models that are advertised as open are not, as only their model weights are published under an open license, but not the training data used to create them. This lack of transparency makes the use of ``open-weight-only'' models risky in science and practice. On the one hand, the effectiveness of a model cannot be verified from scratch on a given benchmark, which carries the risk of inflated scores due to train-test leakage between the model's training data and the test data of the benchmark. On the other hand, using a model with only open weights as part of a productive system carries the risk of copyright infringement, as its training data may contain copyrighted material. Even though the authors of open-weight-only models usually disclaim both, they also typically disclaim any liability for such infringements due to using their model. This typically makes open-weight-only models unusable in industry systems as well as in open source projects.

%%% Contribution.
We therefore introduce the \emph{German Commons}, the largest pre\-training-scale collection of explicitly openly licensed text in German language (Table~\ref{tab:overview}). It encompasses \gcNumTokensInBillions{}~billion tokens of open text across \gcNumDocsInMillions{}~million documents spanning seven thematic domains. This renders it the largest openly licensed German corpus (Section~\ref{related} and~\ref{sources}). Alongside the corpus, we release our data processing library \emph{llmdata} to ensure full reproducibility (Section~\ref{llmdata}). Moreover, we provide an analysis of the corpus' properties (Section~\ref{corpus-statistics}). In the Appendix, a datasheet compliant with the recommendation of~\citet{gebru:2021} is included.

%% file: table-overview.tex
\sffamily
\begin{tabularx}{\linewidth}{
    @{}l
    @{\hspace{4pt}}X
    S[
        table-format=2.2,
        round-mode=places, 
        round-precision=2,
        scientific-notation=fixed, 
        fixed-exponent=6, 
        table-omit-exponent,
        table-omit-exponent,
        table-align-text-post=true,
        table-space-text-post={\,M},
    ]<{\,M} 
    >{\color{gray-600}}S[
        table-format=2.2,
        round-mode=places, 
        round-precision=2,
        scientific-notation=fixed, 
        fixed-exponent=0,
        table-omit-exponent,
        table-align-text-post=true,
        table-space-text-post={\,\%}, 
    ]<{\,\%}
    S[
        table-format=3.2,
        round-mode=places, 
        round-precision=2,
        scientific-notation=fixed, 
        fixed-exponent=9, 
        table-omit-exponent,
        table-align-text-post=true,
        table-space-text-post={\,B},
    ]<{\,B}
    >{\color{gray-600}}S[
        table-format=2.2,
        round-mode=places, 
        round-precision=2,
        scientific-notation=fixed, 
        fixed-exponent=0,
        table-omit-exponent,
        table-align-text-post=true,
        table-space-text-post={\,\%}, 
    ]<{\,\%}@{}
}
\toprule
\multicolumn{6}{c}{\bfseries German Commons} \\
\midrule
\multicolumn{2}{@{}c}{\bfseries Domain} & \multicolumn{2}{c@{}}{\bfseries Documents}    & \multicolumn{2}{c}{\bfseries Tokens}  \\
\cmidrule(r){1-2}                       \cmidrule(lr){3-4}                  \cmidrule(l){5-6} 
\emojiglobe               & Web         &              15476932 & 43.26                 &    19887884828 & 12.87                \\ 
\emojispeech              & Political   &                257888 &  0.72                 &     3565821131 &  2.31                \\ 
\emojiscales              & Legal       &                514726 &  1.44                 &     2992488054 &  1.94                \\ 
\emojinewspaper           & News        &              13266052 & 37.08                 &    72673560680 & 47.02                \\ 
\emojibank                & Economics   &                 57214 &  0.16                 &      110611112 &  0.07                \\ 
\emojibooks               & Cultural    &               6111710 & 17.08                 &    54488679815 & 35.25                \\ 
\emojimicroscope          & Scientific  &                 93689 &  0.26                 &      839151341 &  0.54                \\ 
\midrule
$\sum$ & \bfseries Total  & 35778211              & \multicolumn{1}{c}{}  & 154558196961   & \multicolumn{1}{c}{} \\
\bottomrule
\end{tabularx}

%% file: conf25-german-commons-part2.tex
\section{Related Work}
\label{related}
To motivate our choices for constructing the \emph{German Commons}, we revisit three categories of existing training corpora: \Ni~the web-scraped datasets that dominate current LLM training, \Nii~smaller-scale, German-specific resources, and \Niii~emerging openly licensed alternatives, predominantly in English.

\paragraph{Web Corpora for LLM Training} 

Modern LLM training relies on web-scraped content as large-scale text source, with collections such as \emph{C4}~\cite{raffel:2020}/\emph{mC4}~\cite{xue:2021}, \emph{The~Pile}~\cite{gao:2020}, \emph{OSCAR}~\cite{abadji:2022}, \emph{ROOTS}~\cite{laurencon:2022}, \emph{RedPajama}~\cite{weber:2024}, \emph{Dolma}~\cite{soldaini:2024}, \emph{HPLT}~\cite{de-gibert:2024}, and \emph{FineWeb}~\cite{penedo:2024b,penedo:2025}. However, these datasets derive almost exclusively from Common Crawl\footnote{\url{https://commoncrawl.org/}}, which creates dependency on a single source and lacks explicit licensing metadata. While C4Corpus~\cite{habernal:2016} identifies licensed content through substring matching, license scope remains unclear—content may be \emph{open} but not \emph{verifiable}. Additional risks include terms of service restrictions that prohibit model training~\cite{bommarito:2025} and a considerable amount of PII despite preprocessing~\cite{hong:2025}. The heterogeneous nature of web data further necessitates extensive quality filtering~\cite{soldaini:2024,longpre:2024,wang:2024}. This creates a fundamental tension: web-scraped corpora provide scale but introduce legal, ethical, and quality risks. The \emph{German Commons} aim to address these shortcomings by collecting verifiably licensed, high-quality text content from non-web sources.

\paragraph{German Text Datasets}

Web datasets naturally include German subsets, however, with identical licensing and quality issues. Several additional large-scale German text corpora are available, but mostly predate LLM training efforts, such as the \emph{Leipzig Corpora Collection}~\cite{goldhahn:2012}, \emph{OPUS}~\cite{tiedemann:2012}, and \emph{HPLT}~\cite{de-gibert:2024}. Moreover, since they are also sourced from the web, their licenses are frequently unclear and not verifiable as well. While openly licensed German corpora exist (see Section~\ref{sources}), their individual volumes are substantially smaller than web-scraped alternatives, and they remain fragmented without centralized access. Furthermore, multiple of these datasets are not available in plain text form, and thus need to undergo text extraction and preprocessing first. The \emph{German Commons} aim to instead provide a unified, comprehensive source of text data suitable for large language model training.

\paragraph{Openly Licensed Training Corpora} 

Scaling openly licensed text corpora faces verification challenges, leading datasets to include questionable content~\cite{longpre:2023}. While code datasets can rely on the explicit machine-readable licensing present in code repositories~\cite{kocetkov:2023}, such annotations are rarely available for natural-language text from web or print sources, which consequently proves more difficult to verify. Therefore, data collection efforts turn to trusted providers of licensed and open-domain content, such as government agencies, GLAM institutions, and collaborative projects such as Wikimedia. Current large-scale openly-licensed collections for language model training include:
\Ni~the \emph{Open License Corpus}~\cite{min:2024} is an aggregation from existing corpora covering open domain and attribution licensed content from the legal, scientific, conversational, books, and news domain, amounting to 228B tokens of multi-domain English text;
\Nii~the \emph{KL3M} project~\cite{bommarito:2025} assembled 1.35 trillion tokens of English text sourced primarily from open domain government records in English language; it explicitly excludes content licenses with a `share-alike' clause such as CC-BY-SA and thus removes, e.g., Wikipedia from consideration. 
\Niii~the \emph{Common Pile}~\citep{kandpal:2025} compiles approximately upwards of two trillion tokens of English text with stringent licensing requirements, and is the largest, yet monolingual resource of verifiably open text; and
\Niv~the \emph{Common Corpus}~\cite{langlais:2025} also provides 2 trillion multilingual tokens, including 112 billion German tokens, which serve as a the starting point for our data collection efforts. 

%% file: conf25-german-commons-part3.tex
\section{Sourcing Open German Text Data}
\label{sources}
This section provides our working definition of `openly licensed' and detailed provenance of all data included in the \emph{German Commons}. Dataset construction begins with the German subset of the existing \emph{Common Corpus}~\cite{langlais:2025}, applying stricter licensing and quality criteria that reduce usable tokens from ~112B to ~70B. We then expand coverage through updated source dataset versions and previously unconsidered collections, combining existing resources with newly assembled open data. Table~\ref{tab:data-sources} details all constituent datasets with source, license, type, and size statistics. Data collection uses the newest available versions through August 31st, 2025. 

\begin{table*}[ht]
    \centering
    \caption{Overview on datasets constituting the German Commons corpus. Token counts are measured using GPT-2 tokenizer. `Various' licenses are open, but differ per document as per original source.}
    \input{table-data-sources}
    \label{tab:data-sources}
\end{table*}

\subsection{Where do we obtain data from?}
We identify two source types for \emph{German Commons}: 
\Ni 
established open corpora with published full texts; and 
\Nii 
(metadata) collections requiring text extraction from source files.
We use provided texts where available, otherwise crawling and extracting plain text from sources. Data integration spans multiple providers: Text+, Zenodo, Huggingface, German National Library (DNB), Austrian National Library (ÖNB), German Digital Dictionary (DWDS), Leibniz-Institute for German Language (IDS), and Wikimedia projects.\footnotemark From their catalogs, we obtain all primarily German collections with explicit open licenses. For the non-curated platforms Zenodo and Huggingface, we select only uploads from trusted parties with clear licensing protocols.

\footnotetext{\\[-\baselineskip]
\begin{tabularx}{\linewidth}{Xl}
Text+                                               & \url{https://text-plus.org}               \\
Zenodo                                              & \url{https://zenodo.org}                  \\
Huggingface                                         & \url{https://huggingface.co/datasets}     \\
German National Library (DNB)                       & \url{https://dnb.de}                      \\
Austrian National Library (ÖNB)                     & \url{https://www.onb.ac.at}               \\
German Digital Dictionary (DWDS)                    & \url{https://www.dwds.de/}                \\
Leibniz-Institute for German Language (IDS)         & \url{https://www.ids-mannheim.de}         \\
OpenAlex                                            & \url{https://openalex.org}                \\
Wikimedia                                           & \url{https://www.wikimedia.de}            \\
\end{tabularx}
}

\begin{table}[t]
\centering
\caption{Licenses of data constituting \emph{German Commons}.}
\label{tab:licensing}
\input{table-licensing}
\end{table}

\subsection{What do we consider openly licensed?}

The \emph{German Commons} require explicit licenses for each document. We exclude datasets with ambiguous licensing, i.e., where the aggregation or metadata carries open licenses while underlying text content retains unspecified copyright restrictions. This excludes most web-crawled datasets that conflate aggregation with content licensing rights~\citep{kandpal:2025}. 

We follow \citet{kandpal:2025} in adopting the Open Knowledge Foundation's Open Definition 2.1\footnote{\url{https://opendefinition.org/od/2.1/en/}}. Unlike \emph{Open License Corpus} and \emph{KL3M}, this covers licenses with a share-alike provision. Table~\ref{tab:licensing} lists the accepted licenses found in our data. Licenses are grouped into the categories of 
\Ni public domain equivalent
\Nii attribution licenses; and
\Niii copyleft licenses. 
All of the selected licenses permit redistribution, modification, and commercial use. They thus support data commons principles for sustainable open model development~\cite{lee:2024,tarkowski:2025}. The latter two require attribution and/or license indication, while copyleft licenses have to be redistributed under the same license terms (share-alike). Each document in the \emph{German Commons} is tagged with its corresponding SPDX-canonical license\footnote{\url{https://spdx.org/licenses/}}, linking to its original license text. We exclude licenses with non-commercial clauses, research-only provisions, and other use-limiting conditions. However, we advice practitioners to review individual license compatibility before use.

We acknowledge that this approach, i.e., trusting provided licenses without independent auditing, carries inherent misattribution risks. However, given the relative scarcity of open German text, and the institutional nature of the data providers contributing to the \emph{German Commons}, we consider it a viable method for large-scale data collection. While we apply less strict criteria than \citet{kandpal:2025}, who independently audit entries in their data, to reasonably mitigate risks, we consequently limit inclusion of data to established institutional providers: national libraries, academic institutions, government agencies, and verified open-source platforms, excluding sources lacking clear licensing protocols.

\newpage
\subsection{Detailed Provenance Information}

The \emph{German Commons} are divided into seven thematic domains: \emph{Web} spans collaborative platforms, user-generated content and discussions, and open-source repositories; \emph{Political} encompasses parliamentary protocols, publications, and speeches; \emph{Legal} includes court decisions, proceedings, and regulatory frameworks from German and European judicial systems. While previous datasets group political and legal text as one~\cite{kandpal:2025}, we argue that legal text has a distinct writing style not overlapping with the more general text published by political institutions and thus chose to differentiate between both. The \emph{News} content draws primarily from historical newspaper archives maintained by cultural heritage institutions. \emph{Economics} includes documents from business and trade publications, \emph{Cultural} includes books and general speech transcripts, and \emph{Scientific} includes scholarly and educational content. The constituting data of each domain is detailed in the following paragraphs.

\paragraph{Web Commons} 
For \emph{Wikipedia} and \emph{Wikivoyage}, we use the TEI-encoded version supplied by the DWDS~\citep{nolda:2025a,nolda:2025b}. The complementary corpus \emph{Wikipedia Discussions} of user discussions on Wikipedia is supplied by the IDS~\citep{margaretha:2014}. The \emph{One Million Posts Corpus} consists of user comments posted to the Austrian newspaper `Der Standard', collected and openly licensed in collaboration with the newspaper~\citep{schabus:2017}. 
\emph{Youtube-Commons}~\cite{langlais:2025} encompasses audio transcripts of over 2 million videos shared on YouTube under a CC-BY license. These include both automatically translated and manually curated texts. Finally, we filter German-language website sources and README files hosted on GitHub by filtering the Markdown and TXT subsets of \emph{The Stack}~\citep{kocetkov:2023} corpus for our allowed licenses.

\paragraph{Political Commons}
We aggregate official publications from the German federal parliament (\emph{Corpus der Drucksachen des Deutschen Bundestags}, \cite{10.5281/zenodo.4643066}) and the EU parliament (\emph{EuroVoc}). Alongside that, we include parliamentary protocols and speeches from the German federal parliament (\emph{Corpus der Plenarprotokolle des Deutschen Bundestags}, \cite{10.5281/zenodo.4542662}) and the historic Reichstag (\emph{Reichtagsprotokolle}, \cite{boenig:2023}). German parliamentary documents are exempt from copyright as official works. EU documents are licensed under the European Union Public License (EUPL), which is compatible with a CC BY-SA 3.0 license.\footnote{\url{https://eupl.eu/1.2/en}} Additionally, we include a corpus of political speeches in German language~\cite{barbaresi:2019}.

\paragraph{Legal Commons}
The largest portion of legal documents comprises court proceedings from German courts, collected by \emph{OpenLegalData}~\cite{ostendorff:2020}.\footnote{\url{https://openlegaldata.io}} Proceedings of federal courts are made available in dedicated corpora, namely for the Bundesfinanzhof (\emph{BFH}, \cite{10.5281/zenodo.14622341}), Bundesgerichtshof (\emph{BGH}, \cite{10.5281/zenodo.4540377,10.5281/zenodo.12814022}), Bundesverfassungsgericht (\emph{BVerfG}, \cite{10.5281/zenodo.12705674,10.5281/zenodo.10783177}), Bundespatentgericht (\emph{BPatG}, \cite{10.5281/zenodo.10849977}), Bundesverwaltungsgericht (\emph{BVerwG}, \cite{10.5281/zenodo.10809039}), and Bundesarbeitsgericht (\emph{BAG},  \cite{10.5281/zenodo.4006645}). Court decisions and decrees of German courts are exempt from copyright. Additionally, we include European laws made available via the \emph{EUR-Lex}\footnote{\url{https://eur-lex.europa.eu/}} platform, incorporating the official data dump.

\paragraph{News Commons}
The \emph{Deutsches Zeitungsportal}\footnote{\url{https://www.deutsche-digitale-bibliothek.de/newspaper}} corpus covers over 4 million German newspaper editions of nearly 2000 newspapers from 1671 to 1994, incorporating all public domain releases. The \emph{ANNO} corpus provides equivalent coverage for Austrian newspapers, spanning around 1600 newspapers in the public domain. Additionally, we include the \emph{Europeana Newspapers} archive, which contains more than 4 million individual documents across multiple European languages from the 18th to early 20th century, using the plain text version created by the BigLAM initiative.\footnote{\url{https://huggingface.co/biglam}} \emph{Wikinews} is a wiki project for news articles written by community members; we obtain a current dump and parse plain text from it.

\paragraph{Economics Commons}
The \emph{TEDEUTenders}~\cite{langlais:2025} dataset comprises European public procurement notices from the online version of the EU's Official Journal Supplement, providing structured access to public tender information across EU member states. 

%% OMITTED; this is not included (yet) since DNB has not delivered data 
%The \emph{Börsenblatt für den Deutschen Buchhandel} dataset comprises the digitized historical archive (1834-1945) of the official publication of Germany's book trade association, documenting comprehensive book market developments, industry announcements, and trade information.

\paragraph{Cultural Commons}
The \emph{DiBiLit}~\citep{boenig:2021} dataset provides a literary corpus spanning the 16th through early 20th centuries, containing primarily German literary works (poetry, drama, prose) alongside select humanities texts. \emph{DiBiPhil} covers the 15th to 20th centuries, encompassing literary works with philosophical content. Both corpora feature TEI-encoded, homogenized texts. \emph{Wikisource} is a collaborative digital library for public domain source texts, including historical documents, literary works, and reference materials, transcribed and proofread by volunteer contributors. \emph{GermanPD}~\citep{langlais:2025} systematically aggregates German monographs and periodicals from Internet Archive and European national libraries, including both OCR-sourced content and text extracted from machine-readable format. \emph{BLBooks}~\citep{britishlibrary:2021} comprises digitized pages from the British Library, primarily concentrated on the 18th-19th century and covering diverse subject areas. While BLBooks is originally split into pages, we concatenate pages of the same in the provided order to re-assemble complete texts. \emph{MOSEL}~\citep{gaido:2024} aggregates multilingual open-source speech recordings with Whisper-generated transcriptions across 24 EU languages. \emph{SBB Fulltexts}~\cite{labusch:2023} is a collection of the fulltexts available in the digitized collections of the Berlin State Library (SBB), covering approximately 25\,000 unique German literary works split into individual page scans. \emph{Wikiquote} collaboratively collects quotations and proverbs that fall under public domain.

\paragraph{Scientific Commons}
The \emph{Digitalisierung des Polytechnischen Journals} is a historic technical periodical, digitized through OCR~\citep{hug:2010}. \emph{Wikibooks} is a free repository of open-content textbooks and educational materials, converted to TEI~\citep{nolda:2025c}. We crawl the \emph{Directory of Open Access} books (DOAB) for scholarly open-access book publications and filter for German language, similar to the English counterpart assembled by~\citet{kandpal:2025}. We further include German scholarly articles published on \emph{arXiv} which explicitly indicate open licensing. For each article, we only include the latest version to reduce deduplication overhead. Finally, we obtain all fulltexts from \emph{OpenAlex}~\cite{priem:2022}, a metadata aggregator for open-access scholarly papers, and filter for German texts under open licenses. Since OpenAlex does not provide fulltexts, we instead crawl all linked PDFs and extract plain text from those.

%% file: table-data-sources.tex
\newcommand{\categoryheader}[6]{\\[-8pt] \midrule \multicolumn{5}{@{}l@{}}{\bfseries #2} & #3 & #4 & #5 & #6 \\[-1pt] \cmidrule{6-9}}
\sffamily
\setlength{\tabcolsep}{4pt}
\renewcommand{\arraystretch}{0.75}
\small
\begin{tabularx}{\linewidth}{
    @{}Xl@{}lll@{}
    S[
        table-format=2.2,
        round-mode=places, 
        round-precision=2,
        scientific-notation=fixed, 
        fixed-exponent=6, 
        table-omit-exponent,
        table-omit-exponent,
        table-align-text-post=true,
        table-space-text-post={\,M},
    ]<{\,M}
    >{\color{gray-600}}S[
        table-format=2.2,
        round-mode=places, 
        round-precision=2,
        scientific-notation=fixed, 
        fixed-exponent=0,
        table-omit-exponent,
        table-align-text-post=true,
        table-space-text-post={\,\%}, 
    ]<{\,\%} 
    S[
        table-format=3.2,
        round-mode=places, 
        round-precision=2,
        scientific-notation=fixed, 
        fixed-exponent=9, 
        table-omit-exponent,
        table-align-text-post=true,
        table-space-text-post={\,B},
    ]<{{\,B}}
    >{\color{gray-600}}S[
        table-format=2.2,
        round-mode=places, 
        round-precision=2,
        scientific-notation=fixed, 
        fixed-exponent=0,
        table-omit-exponent,
        table-align-text-post=true,
        table-space-text-post={\,\%}, 
    ]<{\,\%} 
    @{}    
}

\toprule
\bfseries  Thematic Subset \& Constituent Corpora
& \multicolumn{2}{c}{\bfseries Ref.}
& \bfseries License
& \bfseries Text Type
& \multicolumn{2}{c}{\bfseries Docs (\# / \%)}
& \multicolumn{2}{c}{\bfseries Tokens (\# / \%)}\\

\categoryheader{gray-300}{\emojiglobe~Web Commons}{15476932}{43.26}{19887884828}{12.87} 
Wikipedia                                                       & \cite{nolda:2025a}                & \source{https://zenodo.org/records/14748605}                                  & CC-BY-SA-4.0      & Various                    & 2930224 &  8.19 &  2948751608 &  1.91 \\
Wikivoyage                                                      & \cite{nolda:2025b}                & \source{https://zenodo.org/records/14748553}                                  & CC-BY-SA-4.0      & Travel                     &   20370 &  0.06 &    42025478 &  0.03 \\ 
Wikipedia Discussions                                           & \cite{margaretha:2014}            & \source{https://corpora.ids-mannheim.de/pub/wikipedia-deutsch/2024/}          & CC-BY-SA-4.0      & Online Discussions         & 8349076 & 23.34 &  1218210917 &  0.79 \\
Youtube-Commons                                                 & \cite{langlais:2025}              & \source{https://huggingface.co/datasets/PleIAs/YouTube-Commons}               & Various           & Video Subtitles            & 2809714 &  7.85 & 14478850964 &  9.37 \\ 
One Million Posts Corpus                                        & \cite{schabus:2017}               & \source{https://ofai.github.io/million-post-corpus/}                          & CC-BY-4.0         & Online Discussions         &  946082 &  2.64 &    94872633 &  0.06 \\
The Stack (Markdown and TXT Subsets)                            & \cite{kocetkov:2023}              & \source{https://huggingface.co/datasets/bigcode/the-stack-dedup}              & Various           & Various                    &  421466 &  1.18 &  1105173228 &  0.72 \\
  
\categoryheader{gray-300}{\emojispeech~Political Commons}{257888}{0.72}{3565821131}{2.31} 
Reichtagsprotokolle                                             & \cite{boenig:2023}                & \source{https://zenodo.org/records/10225467}                                  & CC-BY-SA-4.0      & Parliamentary Protocols    &     522 &  0.00 &   703495637 &  0.46 \\
German Political Speeches                                       & \cite{barbaresi:2019}             & \source{ttps://zenodo.org/records/3611246}                                    & CC-BY-4.0         & Speech Transcripts         &    6678 &  0.02 &    29409655 &  0.02 \\
Corpus der Drucksachen des Deutschen Bundestages                & \cite{10.5281/zenodo.4643066}     & \source{https://zenodo.org/records/4643066}                                   & CC0-1.0           & Parliamentary Publications &    3017 &  0.01 &   528769669 &  0.34 \\ 
C. d. Plenarprotokolle des Deutschen Bundestages           & \cite{10.5281/zenodo.4542662}     & \source{https://zenodo.org/records/4542662}                                   & CC0-1.0           & Parliamentary Protocols    &    1833 &  0.01 &   316034708 &  0.20 \\
EuroVoc                                                         &                                   & \source{https://huggingface.co/datasets/EuropeanParliament/Eurovoc}           & EUPL              & Parliamentary Publications &  245838 &  0.69 &  1988111462 &  1.29 \\

\categoryheader{gray-300}{\emojiscales~Legal Commons}{514726}{1.44}{2992488054}{1.94} 
Corpus des Deutschen Bundesrechts                               & \cite{10.5281/zenodo.14592346}    & \source{https://zenodo.org/records/14592346}                                  & CC0-1.0           & German Federal Laws        &    3217 &  0.01 &     1004294 &  0.00 \\
OpenLegalData                                                   & \cite{ostendorff:2020}            & \source{https://huggingface.co/datasets/schneiderai/openlegaldata}            & CC0-1.0           & Court Decisions            &  227955 &  0.70 &  1915956613 &  1.24 \\
Corpus der Entscheidungen des BFH                               & \cite{10.5281/zenodo.14622341}    & \source{https://zenodo.org/records/14622341}                                  & CC0-1.0           & Court Decisions            &   10885 &  0.03 &    67791931 &  0.04 \\
Entscheidungen des BGH in Strafsachen des 20. Jhd.              & \cite{10.5281/zenodo.4540377}     & \source{https://zenodo.org/records/4540377}                                   & CC0-1.0           & Court Decisions            &   36062 &  0.10 &    92873390 &  0.06 \\
Corpus der Entscheidungen des BGH                               & \cite{10.5281/zenodo.12814022}    & \source{https://zenodo.org/records/12814022}                                  & CC0-1.0           & Court Decisions            &   77258 &  0.22 &   292832709 &  0.19 \\
Corpus der Entscheidungen des BVerfG                            & \cite{10.5281/zenodo.12705674}    & \source{https://zenodo.org/records/12705674}                                  & CC0-1.0           & Court Decisions            &    8028 &  0.02 &    39503223 &  0.03 \\
Corpus der Entscheidungen des BpatG                             & \cite{10.5281/zenodo.10849977}    & \source{https://zenodo.org/records/10849977}                                  & CC0-1.0           & Court Decisions            &   30705 &  0.09 &   185099188 &  0.12 \\
Corpus der Entscheidungen des BVerwG                            & \cite{10.5281/zenodo.10809039}    & \source{https://zenodo.org/records/10809039}                                  & CC0-1.0           & Court Decisions            &   27185 &  0.08 &   123487739 &  0.08 \\
Corpus der amtl. Entscheidungssammlung des BVerfG               & \cite{10.5281/zenodo.10783177}    & \source{https://zenodo.org/records/10783177}                                  & CC0-1.0           & Court Decisions            &     919 &  0.00 &    24427294 &  0.02 \\
Corpus der Entscheidungen des BAG                               & \cite{10.5281/zenodo.4006645}     & \source{https://zenodo.org/records/4006645}                                   & CC0-1.0           & Court Decisions            &    5624 &  0.02 &    48248111 &  0.03 \\
EurLEX                                                          & \cite{chalkidis:2021}             & \source{https://zenodo.org/record/5363165/}                                   & CC-BY-4.0         & European Union Laws        &   64934 &  0.18 &   201263562 &  0.13 \\

\categoryheader{gray-300}{\emojinewspaper~News Commons}{13266052}{37.08}{72673560680}{47.02} 
Deutsches Zeitungsportal                                        &                                   & \source{https://www.deutsche-digitale-bibliothek.de/newspaper}                & CC0-1.0           & News Articles              & 8076164 & 22.57 & 43871094547 & 28.38 \\ 
Europeana Newspapers                                            &                                   & \source{https://huggingface.co/datasets/biglam/europeana_newspapers}          & CC0-1.0           & News Articles              & 3256341 &  9.10 & 20684418365 & 13.38 \\ 
ANNO                                                            &                                   & \source{https://labs.onb.ac.at/en/datasets/anno/}                             & CC0-1.0           & News Articles              & 1910281 &  5.34 &  8103825248 &  5.24 \\
Wikinews                                                        &                                   & \source{https://dumps.wikimedia.org/dewikinews/20250820}                      & CC-BY-SA-4.0      & News Articles              &   23266 &  0.07 &    14222520 &  0.01 \\

\categoryheader{gray-300}{\emojibank~Economics Commons}{57214}{0.16}{110611112}{0.07} 
TEDEUTenders                                                    & \cite{langlais:2025}              & \source{https://huggingface.co/datasets/PleIAs/TEDEUTenders}                  & CC0-1.0           & Procurement Notices        &   57214 &  0.16 &   110611112 &  0.07 \\

\categoryheader{gray-300}{\emojibooks~Cultural Commons}{6111710}{17.08}{54488679815}{35.25} 
DiBiLit-Korpus                                                  & \cite{boenig:2021}                & \source{https://zenodo.org/records/5786725}                                   & CC-BY-SA-4.0      & Literature                 &    2062 &  0.01 &   216391448 &  0.14 \\
DiBiPhil-Korpus                                                 &                                   & \source{https://github.com/deutschestextarchiv/DiBiPhil}                      & CC-BY-SA-4.0      & Literature                 &     269 &  0.00 &    32151997 &  0.02 \\
Wikisource                                                      &                                   & \source{https://dumps.wikimedia.org/dewikisource/20250801/}                   & CC-BY-SA-4.0      & Various                    &  240689 &  0.67 &   347770430 &  0.23 \\
German-PD                                                       & \cite{langlais:2025}              & \source{https://huggingface.co/datasets/PleIAs/German-PD}                     & CC0-1.0           & Literature                 &  123592 &  0.35 & 49333198231 & 31.92 \\
BLBooks                                                         & \cite{britishlibrary:2021}        & \source{https://huggingface.co/datasets/biglam/blbooks-parquet}               & CC0-1.0           & Literature                 &    3714 &  0.01 &  1012047216 &  0.65 \\
MOSEL                                                           & \cite{gaido:2024}                 & \source{https://huggingface.co/datasets/FBK-MT/mosel}                         & CC-BY-4.0         & Speech Transcripts         & 3127203 &  8.74 &  3181917752 &  2.06 \\
SBB Fulltexts                                                   & \cite{labusch:2023}               & \source{https://zenodo.org/records/7716098}                                   & CC-BY-4.0         & Literature                 & 2605569 &  7.28 &   358514283 &  0.23 \\
Wikiquote                                                       &                                   & \source{https://dumps.wikimedia.org/dewikiquote/20250820}                     & CC-BY-SA-4.0      & Quotes \& Proverbs         &    8612 &  0.02 &     6688458 &  0.00 \\
  
\categoryheader{gray-300}{\emojimicroscope~Scientific Commons}{93689}{0.26}{839151341}{0.54} 
Wikibooks                                                       & \cite{nolda:2025c}                & \source{https://zenodo.org/records/14748586}                                  & CC-BY-SA-4.0      & Educational Books          &   27292 &  0.08 &    50434996 & 0.03 \\
Digitalisierung des Polytechnischen Journals                    & \cite{hug:2010}                   & \source{https://github.com/deutschestextarchiv/dingler}                       & CC-BY-SA-4.0      & Scholarly Papers           &     346 &  0.00 &   180257799 & 0.12 \\
Directory of Open Access Books                                  &                                   & \source{https://www.doabooks.org}                                             & Various           & Scholarly Books            &    1939 &  0.01 &   166920321 & 0.11 \\
arXiv                                                           &                                   & \source{https://www.kaggle.com/datasets/Cornell-University/arxiv}             & Various           & Scholarly Papers           &       8 &  0.00 &      103478 & 0.00 \\
Wikiversity                                                     &                                   & \source{https://dumps.wikimedia.org/dewikiversity/20250820}                   & CC-BY-SA-4.0      & Educational Content        &   16371 &  0.05 &    27802099 & 0.02 \\
OpenALEX                                                        & \cite{priem:2022}                 & \source{https://openalex.org/}                                                & Various           & Scholarly Papers           &   47733 &  0.13 &   413632648 & 0.27 \\

\midrule
\multicolumn{5}{@{}l@{}}{\bfseries Total}     
&     35778211 & \multicolumn{1}{c}{}  
& 154558196961 & \multicolumn{1}{c}{}   \\
\bottomrule
\end{tabularx}

%% file: table-licensing.tex
\small
\sffamily
\setlength{\tabcolsep}{3pt}
\begin{tabularx}{\linewidth}{@{}lll*{4}{c}@{}}
\toprule
\multicolumn{3}{l}{\textbf{License}}                                           &
\multicolumn{1}{r}{\lapbox[.8cm]{0cm}{\rotatebox[origin=rB]{-25}{\textbf{Attribution}}}}        & 
\multicolumn{1}{r}{\lapbox[.8cm]{0cm}{\rotatebox[origin=rB]{-25}{\textbf{Copyright Notice}}}}   & 
\multicolumn{1}{r}{\lapbox[.8cm]{0cm}{\rotatebox[origin=rB]{-25}{\textbf{Share-Alike}}}}        & 
\multicolumn{1}{r}{\lapbox[.8cm]{0cm}{\rotatebox[origin=rB]{-25}{\textbf{Source Provision}}}}   \\
&                         &                                                               & $\downarrow$  & $\downarrow$  & $\downarrow$  & $\downarrow$  \\ \midrule
\multirow{4}{*}{\rotatebox[origin=c]{90}{PD-Equiv.}}
& CC0-1.0 / Public Domain & \source{https://spdx.org/licenses/CC0-1.0.html}               &               &               &               &               \\
& Unlicense               & \source{https://spdx.org/licenses/Unlicense.html}             &               &               &               &               \\
& MIT-0                   & \source{https://spdx.org/licenses/MIT-0.html}                 &               &               &               &               \\
& 0BSD                    & \source{https://spdx.org/licenses/0BSD.html}                  &               &               &               &               \\\midrule

\multirow{12}{*}{\rotatebox[origin=c]{90}{Attribution}}
& MIT                     & \source{https://spdx.org/licenses/MIT.html}                   & \checkmark    & \checkmark    &               &               \\
& BSD-2-Clause            & \source{https://spdx.org/licenses/BSD-2-Clause.html}          & \checkmark    & \checkmark    &               &               \\
& BSD-2-Clause-FreeBSD    & \source{https://spdx.org/licenses/BSD-2-Clause-FreeBSD.html}  & \checkmark    & \checkmark    &               &               \\
& BSD-3-Clause            & \source{https://spdx.org/licenses/BSD-3-Clause.html}          & \checkmark    & \checkmark    &               &               \\
& BSD-Source-Code         & \source{https://spdx.org/licenses/BSD-Source-Code.html}       & \checkmark    & \checkmark    &               &               \\
& Apache-1.1              & \source{https://spdx.org/licenses/Apache-1.1.html}            & \checkmark    & \checkmark    &               &               \\
& Apache-2.0              & \source{https://spdx.org/licenses/Apache-2.0.html}            & \checkmark    & \checkmark    &               &               \\
& BSD-4-Clause-UC         & \source{https://spdx.org/licenses/BSD-4-Clause-UC.html}       & \checkmark    & \checkmark    &               &               \\
& BSD-4-Clause            & \source{https://spdx.org/licenses/BSD-4-Clause.html}          & \checkmark    & \checkmark    &               &               \\
& CC-BY-2.0               & \source{https://spdx.org/licenses/CC-BY-2.0.html}             & \checkmark    & \checkmark    &               &               \\
& CC-BY-3.0               & \source{https://spdx.org/licenses/CC-BY-3.0.html}             & \checkmark    & \checkmark    &               &               \\
& CC-BY-4.0               & \source{https://spdx.org/licenses/CC-BY-4.0.html}             & \checkmark    & \checkmark    &               &               \\\midrule

\multirow{3}{*}{\rotatebox[origin=c]{90}{Copyleft}}
& CC-BY-SA-4.0            & \source{https://spdx.org/licenses/CC-BY-SA-4.0.html}          & \checkmark    & \checkmark    & \checkmark    &               \\
& EUPL-1.2                & \source{https://spdx.org/licenses/EUPL-1.2.html}              & \checkmark    & \checkmark    & \checkmark    & \checkmark    \\
& Artistic-2.0            & \source{https://spdx.org/licenses/Artistic-2.0.html}          & \checkmark    & \checkmark    & \checkmark    & \checkmark    \\\bottomrule
\end{tabularx}

%% file: conf25-german-commons-part4.tex
\section{Data Processing}
\label{llmdata}

We apply several processing steps to transform the heterogeneous input data into a consistent output format, apply quality heuristics and filters, and fix text formatting. The individual steps are detailed in this section and are executed in the listed order. We include detailed statistics on removed documents and tokens per processing step and source dataset in Appendix~\ref{app:data-filtering-steps}. Table~\ref{tab:schema} shows the final dataset schema. %All code used to assemble the corpus is made available open source.\footnote{\url{https://github.com/coral-nlp/german-commons}} We also make available a scalable pipeline for processing LLM data based on the Ray~\citep{ray:2025} parallel processing framework, allowing third parties to easily recreate, extend, or assemble new data.\footnote{\url{https://github.com/coral-nlp/llmdata}} 
The dataset is distributed in Parquet format, with partial files partitioned for each thematic domain and source dataset to allow selective loading.

\paragraph{Plain Text Extraction}
Most of the compiled source data sets already feature readily available plain text. In instances where the original corpus is only available in PDF format, we use \texttt{Grobid}~\cite{grobid:2025} (for scholarly sources) or \texttt{OlmOCR}~\cite{poznanski:2025} (for other types of PDF) to obtain plain text. In sources where TEI or similar formats featuring structural markup are used, we convert to plain text and systematically exclude editorial elements such as title pages, page numbering and breaks, footnotes, bibliographic references, and textual apparatus, preserving only the core textual content. Markdown syntax is left intact when encountered. For wiki markup, we use \texttt{mwparserfromhell}\footnote{\url{https://github.com/earwig/mwparserfromhell}} to convert to plain text.

\paragraph{Text Formatting}
To address artifacts introduced during optical character recognition present in a large portion of the data, we further apply a minimal amount of text formatting. We apply the standard formatters of the FTFY suite~\cite{speer:2019}, including UTF-8 encoding fixes, removal of HTML entities, control characters and terminal escape sequences, ligature decomposition, character width and surrogate pair fixes, quote character normalization, and Unicode NFC normalization. In addition, we apply a series of custom regex-based transformations to address common whitespace issues resulting from OCR-sourced data. It collapses multiple spaces and tabs into single spaces, reduces excessive newlines while preserving paragraph boundaries, removes hyphenation on line breaks, and removes leading and trailing whitespace from individual lines. 

\paragraph{Language \& Length Filtering}
For datasets that indicate language in their metadata, we pre-filter using each dataset's own language tags to reduce redundant classification work. Then, we employ the FastText language identification model~\cite{joulin:2016} to automatically detect the language of the input texts. We apply the compressed model version~\cite{joulin:2017}, which supports 176 languages, to text snippets truncated to 4096 characters for computational efficiency. The classifier treats the input by replacing newlines with spaces to improve the prediction accuracy, as the FastText model was trained on single-line text samples. We discard text not classified as primarily German with a probability of at least 0.65.

We use the GPT-2 tokenizer~\cite{radford:2019} to obtain token counts for all sequences in the corpus. We discard sequences shorter than 32 tokens, as the majority of those are extraction artifacts. The token count is persisted alongside the text data for downstream filtering.

\begin{table}[tb]
    \centering
    \caption{Dataset schema}
    \label{tab:schema}
    \input{table-schema}
\end{table}

\paragraph{Quality Filtering}
Following the consideration of other large-scale pretraining datasets, such as the BigScience ROOTS corpus~\cite{laurencon:2022} for BLOOM~\cite{scao:2022}, Gopher~\cite{rae:2021}, and CulturaX~\cite{nguyen:2024}, we implement several quality indicators to remove low-quality text. Those include word count, average word length, symbol-to-word ratios (hash and ellipsis), bullet/ellipsis line ratios, alphabetic word ratio, stop word count, text repetition through duplicate line/paragraph fractions, character-level duplicate fractions, and n-gram repetition analysis (2-4 grams for top frequency, 5-10 grams for duplication). We select the exclusion parameters for these indicators using percentile-based thresholds~\cite{nguyen:2024}, calculating the value distributions on language-filtered and formatted data, and removing documents either below the 5th or above the 95th percentile, depending on indicator. As large subset of our data originates from OCR text, we additionally apply OCR-specific filtering heuristics to exclude errors not previously addressed through formatting fixes, namely character casing anomalies, fragmented words, and special character density. Exact parameters for all applied filters can be found in Appendix~\ref{app:data-filtering-parameters}. For all parameters, their choice was manually validated by cursory inspection of removed content for different percentile thresholds. Deviations from parameters used in prior work for English web-text~\cite{rae:2021} are little, and reasonable given the difference in language and domains of our data. 

\paragraph{Deduplication} 
For deduplication, we rely on the LSH bloom filter implementation of Dolma~\cite{soldaini:2024}. We perform paragraph-level deduplication, splitting each document into chunks at newline characters. MinHash collisions are detected using 20-gram shingling and a collision rate of 0.8, i.e., for two paragraphs to be deemed duplicates, 80\% of their ngrams have to be identical. All but one chunk are removed from the corpus in case of collisions. The bloom filter was parametrized for a false positive rate of 1e-4. We make available the bloom filter file for others to deduplicate new data against \emph{German Commons}.

\paragraph{PII Removal}
We remove personally identifiable information (PII) using a combination of regex-based filters and the Presidio framework~\cite{mendels:2018}. Types of PII removed are email addresses, phone numbers, IP addresses, credit-card numbers, IBAN numbers, and URL. To keep the semantic structure of sentences intact, we replace each of them with respective generic information. A list of the replacements strings used is included in Appendix~\ref{app:data-filtering-pii}, which allows lookup for full redaction or other replacements downstream. 

\paragraph{License Mapping and Filtering}
We map the diverse license identifiers indicated in the data to a canonical set of SPDX license URLs pointing to the corresponding license of each document in the corpus. We then filter licenses to only open licenses as listed in Table~\ref{tab:licensing}. For cases where multiple licenses are given for a document, we require all of them to be permissive.

%% file: table-schema.tex
\sffamily
\small
\renewcommand{\arraystretch}{0.85}
\begin{tabularx}{\linewidth}{@{}lX@{}}
\toprule
\bfseries Key           & \bfseries Value \\
\midrule
\texttt{id}             & Document identifier as found in the original dataset          \\
\texttt{source}         & Source dataset this document stems from                       \\
\texttt{subset}         & Thematic subset of this document                              \\
\texttt{text}           & Cleaned full text of this document                            \\
\texttt{license}        & Canonical SPDX license URL(s) of this document                \\
\texttt{num\_tokens}    & Number of GPT-2 tokens in this document                       \\
\texttt{perplexity}     & Perplexity of this document estimated by Wikipedia KenLM      \\
\texttt{ocr\_score}     & OCR quality as calculated by \href{https://github.com/Pleias/OCRoscope}{\texttt{ocroscope}}  \\
\bottomrule
\end{tabularx}

%% file: conf25-german-commons-part5.tex
\section{Corpus Statistics}
\label{corpus-statistics}

We analyze corpus composition across thematic subsets and license types, demonstrate the efficacy of our filtering pipeline, and investigate different text properties, in order to illustrate \emph{German Commons}' suitability for language model pretraining.

\paragraph{Token and Document Distribution By Thematic Domain.}

\begin{table}[t]
\centering
\setlength{\tabcolsep}{7pt}
\small
\caption{Split Composition by domain for different training context lengths $s$. Assumes disjoint splits, i.e., $s \leq 8192$ contains all sequences $2048 < s \leq 8192$. Percentage for $\sum$ is in relation to full dataset.}
\label{tab:tokens-by-subset}
\input{table-tokens-by-subset}
\end{table}

\begin{figure}[t]
    \centering
    \includegraphics[width=\linewidth, trim={2mm 1mm 0mm 2mm}, clip, viewport=0 0 8.48cm 5.3cm]{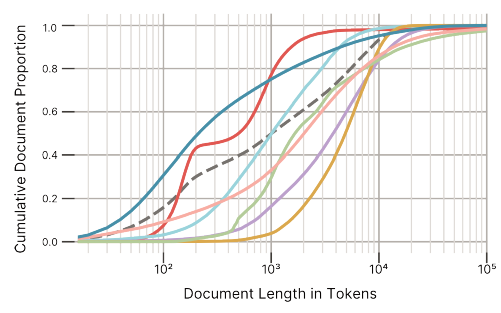}
    \caption{Cumulative proportion of tokens by document length in corpus, normalized by domain;
        \colordot{gray-600}{gray-800} overall (dashed), and by subset for
        \colordot{red-600}{red-800} cultural,
        \colordot{blue-300}{blue-500} economic,
        \colordot{violet-400}{violet-600} legal,
        \colordot{yellow-400}{yellow-600} news,
        \colordot{green-400}{green-600} political,
        \colordot{red-300}{red-500} scientific,
        \colordot{blue-600}{blue-800} web.
    }
    \label{fig:docs-by-subset}
\end{figure}

\Cref{fig:docs-by-subset} shows cumulative document proportions by length across thematic domains in \emph{German Commons}. Distinct length distributions are apparent: cultural and web content concentrate in shorter documents, with cultural content showing rapid cumulative growth below 1,000 tokens due to page-level book segmentation in some source corpora. News content exhibits substantially longer documents, while legal, scientific, and political domains occupy intermediate positions similar to the overall corpus trend. \Cref{tab:tokens-by-subset} quantifies the practical implications of these length distributions for language model training contexts. When partitioning documents into subsets of short ($s \leq 2048$), medium ($s \leq 8192$), long ($s \leq 32\,768$) and very long ($s > 32\,768$) context lengths, domain distribution varies across partitions. At short contexts, web content dominates with 43.31\% of available tokens, followed by news (27.05\%) and cultural content (26.47\%). For medium and long contexts, news articles provide the majority of tokens at 78.60\% and 63.38\% respectively, with web content ranging from to 16.95\% to 31.18\%. At very long contexts, most tokens (83.73\%) originate from book-length documents of the cultural domain. Scientific, legal, and political domains maintain smaller representation across all context lengths, ranging from 0.5-3\% each. \textbf{All thematic domains remain represented in each length partition, enabling sub- or oversampling at different context lengths to control model exposure toward text genres.}

\paragraph{Token and Document Distribution By License Type.}

\begin{table}[t]
\centering
\caption{Number of documents and tokens per license type.}
\label{tab:tokens-per-license}
\input{table-tokens-by-license}
\end{table}

\begin{table}[t]
\centering
\caption{Number of tokens per license type and domain.}
\label{tab:tokens-per-lincense-and-subset}
\input{table-tokens-by-license-and-subset}
\end{table}

\Cref{tab:tokens-per-license} presents document counts and token volume across license categories. Public domain equivalent licenses dominate token count with 126.61B tokens (74.291\% of corpus) despite representing only 35.96\% of documents, reflecting substantially longer average sequence lengths. This length disparity stems primarily from news articles and cultural content under public domain licensing. Attribution-type licenses contribute 34.48B tokens (20.40\% of corpus) across 33.4\% of documents, while copyleft licenses provide 7.93B tokens (4.69\% of corpus) from 30.60\% of documents. With licenses being derived from source datasets, license types exhibit strong correlations with thematic domains. Public domain content concentrates heavily in cultural (39.36\% of public domain tokens) and news domains (57.39\%), with minimal representation in web content (0.08\%). Attribution licenses are predominantly found in web content (88.76\% of attribution tokens), followed by cultural content (10.27\%). Copyleft licenses span web sources (52.29\%, including share-alike licensed Wikimedia projects) and political text (33.84\%, primarily from EUPL-licensed \emph{Eurovoc}). \textbf{All thematic domains feature public domain data, and public domain data yields 75\% of tokens.}

\paragraph{Filtering Statistics.} 
Our data filtering pipeline removes noisy data in three sequential stages  (also see \Cref{app:data-filtering-steps}). 
\Ni 
Quality filtering removed 46.41\% of initial data, with the majority of that being non-German text eliminated from multilingual source corpora (e.g., \emph{The Stack}, \emph{arXiv}) and very short texts being removed (e.g., Wikipedia redirect pages and failed PDF extractions). 
\Nii 
Deduplication only removes an additionaly 2.7\% of text, concentrated in web and news corpora which exhibit both within-corpus and cross-corpus overlaps. Other domains showed minimal duplication.
\Niii 
Final license compliance and PII filtering removed negligible volumes (0.08\%). Source corpora contained minimal personally identifiable information, occurring predominantly in web sources and economics content, while other domains required virtually no PII filtering. Overall retention reached 50.73\% of input, which, however, is attributable to select multilingual corpora, with the majority of source corpora retaining between 70\% and 95\% of their content. \textbf{This aligns with our filtering objective of eliminating noise while maximizing text retention at consistent quality. Only trace amounts of duplicates and PII were present in source corpora and subsequently removed.}

\paragraph{Text Properties.}
We employ four pretrained encoder-based German text classifiers\footnotemark to assess text properties across on a stratified random sample of 10\,000 paragraphs per source corpus (385\,467 total, due to less paragraphs available in some sources).

\footnotetext{
\hbox to 6.5em{Toxicity \cite{arnett:2024}:} \url{https://hf.co/PleIAs/celadon} \\
\phantom{\textsuperscript{11}}\hbox to 6.5em{Sentiment \cite{guhr:2020}:} \url{https://hf.co/oliverguhr/german-sentiment-bert} \\
\phantom{\textsuperscript{11}}\hbox to 6.5em{Complexity:} \url{https://hf.co/krupper/text-complexity-classification} \\
\phantom{\textsuperscript{11}}\hbox to 6.5em{Fluency:} \url{https://hf.co/EIStakovskii/bert-base-german-cased_fluency}
}

\begin{table}[t]
    \caption{Proportion of toxicity degree and corresponding toxicity label across paragraph sample. Differences for domains are minimal. No values above 3 are observed.}
    \label{tab:toxicity}
    \centering
    \input{table-toxicity}
\end{table}
The toxicity classifier grades content on 0-7 scales across five categories (ability, gender/sex, race/origin, religion, and violence), with scores 0-3 considered non-toxic. Results listed in \Cref{tab:toxicity} show no paragraphs scoring above 3 in any category, with on average 95\% of paragraphs scoring 0. Remaining scores fall between 1-3, with negligible differences between thematic domains. \textbf{The \emph{German Commons} is thus deemed to contain only minimal amounts of harmful or toxic content.}

\begin{figure}
    \centering
    \includegraphics[width=\linewidth,trim={2mm 1mm 0mm 0mm},clip, viewport=0 0 8.49cm 3.65cm]{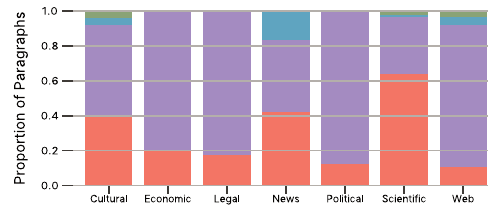}
    \caption{Proportion of text complexity (
        \colordot{green-500}{green-700} easy,
        \colordot{blue-500}{blue-700} simple,
        \colordot{violet-500}{violet-700} everyday,
        \colordot{red-500}{red-700} special
    ) across paragraph sample, per subset.}
    \label{fig:complexity}
\end{figure}

Language complexity classification (\Cref{fig:complexity}) identifies four grades: plain (2\%), easy (3\%), everyday (65\%), and special language (30\%). \Cref{fig:complexity} reveals expected domain variations: scientific content exhibits highest special language proportion (63.8\%), while web content shows highest everyday language (81.4\%), with similar distributions found in Political, Economic, and Legal domains. News content demonstrates intermediate complexity with balanced distribution across categories. Cultural is nearly evenly divided between everyday (52.2\%) and special language (39.9\%). \textbf{Overall, a balanced complexity distribution across domains enables learning across linguistic registers.}

\begin{figure}
    \centering
    \includegraphics[width=\linewidth, trim={2mm 1mm 0mm 0mm}, clip, viewport=0 0 8.49cm 3.65cm]{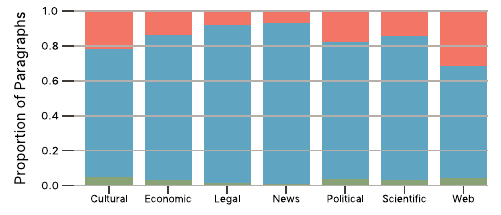}
    \caption{Proportion of text sentiment classes (
        \colordot{red-500}{red-700} negative,
        \colordot{blue-500}{blue-700} neutral,
        \colordot{green-500}{green-700} positive
     ) across paragraph sample, per domain. 
    }
    \label{fig:sentiment}
\end{figure}

Sentiment analysis (\Cref{fig:sentiment}) categorizes text as negative (16.4\%), neutral (80.5\%), or positive (3.1\%). Web content exhibits highest negative sentiment (31.8\%), while cultural content shows most positive sentiment (5.0\%). News content demonstrates the highest proportion of neutral sentiment (92.0\%). The remaining domains maintain proportions similar to the overall distribution. \textbf{Mostly neutral text prevents systematic model biases w.r.t. sentiment.}

%% file: table-tokens-by-subset.tex
\sffamily
\setlength{\tabcolsep}{3pt}
\begin{tabularx}{\linewidth}{
    @{}X
    *{4}{
        S[
            table-format=2.2,
            round-mode=places, 
            round-precision=2,
            scientific-notation=fixed, 
            fixed-exponent=9, 
            table-omit-exponent,
            table-align-text-post=true,
            table-space-text-post={\,B},
        ]<{\,B}@{\hskip 4pt}
        >{\color{gray-600}}S[
            table-format=2.2,
            round-mode=places, 
            round-precision=2,
            scientific-notation=fixed, 
            fixed-exponent=0,
            table-omit-exponent,
            table-align-text-post=true,
            table-space-text-post={\,\%}
        ]<{\,\%}
    }
    @{}
}
\toprule
                    & \multicolumn{2}{c}{$s \leq 2048$} & \multicolumn{2}{c}{$s \leq 8192$} & \multicolumn{2}{c}{$s \leq 32768$} & \multicolumn{2}{c}{$s > 32768$} \\
                    \cmidrule(lr){2-3}                   \cmidrule(lr){4-5}                   \cmidrule(lr){6-7}                    \cmidrule(l){8-9}
\emojibooks         &                3195740496 & 26.47 &                 681460772 &  1.37 &                  401967430 &  0.84 &             49730480107 & 83.72 \\
\emojibank          &                  31650561 &  0.26 &                  57646353 &  0.12 &                   15091680 &  0.03 &                 5890147 &  0.01 \\
\emojiscales        &                 173536111 &  1.44 &                1077371912 &  2.16 &                 1484936172 &  3.11 &               243052235 &  0.41 \\
\emojinewspaper     &                3266246851 & 27.05 &               39166614358 & 78.60 &                30238952090 & 63.38 &                 1382184 &  0.00 \\
\emojispeech        &                 141459420 &  1.17 &                 272344617 &  0.55 &                  505242761 &  1.06 &              2635844020 &  4.44 \\
\emojimicroscope    &                  36360922 &  0.30 &                 126419846 &  0.25 &                  186831387 &  0.39 &               485500106 &  0.82 \\
\emojiglobe         &                5228855902 & 43.31 &                8447950761 & 16.95 &                14876195284 & 31.18 &              6300433630 & 10.61 \\ \midrule
$\sum$              &               12073850263 &  7.14 &               49829808619 & 29.48 &                47709216804 & 28.23 &             59402582429 & 35.15 \\
\bottomrule
\end{tabularx}

%% file: table-tokens-by-license.tex
\sffamily
\setlength{\tabcolsep}{6pt}
\begin{tabularx}{\linewidth}{
    @{}X
    S[
        table-format=2.2,
        tight-spacing=true,
        round-mode=places, 
        round-precision=2,
        scientific-notation=fixed, 
        fixed-exponent=6, 
        table-omit-exponent,
        table-align-text-post=true,
        table-space-text-post={\,M}
    ]<{\,M}
    >{\color{gray-600}}S[
        table-format=2.2,
        tight-spacing=true,
        table-align-text-post=true,
        table-space-text-post={\,\%},
        round-mode=places, 
        round-precision=2,
        scientific-notation=fixed, 
        fixed-exponent=0, 
    ]<{\,\%}
    S[
        table-format=3.2,
        tight-spacing=true,
        round-mode=places, 
        round-precision=2,
        scientific-notation=fixed, 
        fixed-exponent=9, 
        table-omit-exponent,
        table-align-text-post=true,
        table-space-text-post={\,B}
    ]<{\,B}
    >{\color{gray-600}}S[
        table-format=2.2,
        tight-spacing=true,
        table-align-text-post=true,
        table-space-text-post={\,\%},
        round-mode=places, 
        round-precision=2,
        scientific-notation=fixed, 
        fixed-exponent=0, 
    ]<{\,\%}
    @{}
}
\toprule
\bfseries License Type  & \multicolumn{2}{c}{\bfseries Documents} & \multicolumn{2}{c}{\bfseries Tokens} \\
\cmidrule(r){1-1}              \cmidrule(lr){2-3}  \cmidrule(l){4-5}
Public Domain &  13945468 & 35.96 & 126606283770 & 74.91 \\
Attribution   &  12969822 & 33.44 &  34477060097 & 20.40 \\
Copyleft      &  11867621 & 30.60 &   7932114248 &  4.69 \\
\bottomrule
\end{tabularx}

%% file: table-tokens-by-license-and-subset.tex
\sffamily
\setlength{\tabcolsep}{3pt}
\begin{tabularx}{\linewidth}{
    @{}l@{\hskip 3pt}X
    *{3}{
        S[
            tight-spacing=true,
            table-format=2.2,
            round-mode=places, 
            round-precision=2,
            scientific-notation=fixed, 
            fixed-exponent=9, 
            table-omit-exponent,
            table-align-text-post=true,
            table-space-text-post={\,B}
        ]<{\,B}@{\hskip 2.5pt}
        >{\color{gray-600}}S[
            tight-spacing=true,
            table-format=2.2,
            round-mode=places, 
            round-precision=2,
            scientific-notation=fixed, 
            fixed-exponent=0,
            table-align-text-post=true,
            table-space-text-post={\,\%}, 
        ]<{\,\%}
    }
    @{}
}
\toprule
\multicolumn{2}{c}{\bfseries Subset}    & \multicolumn{2}{c}{\bfseries Public-Domain} & \multicolumn{2}{c}{\bfseries Attribution} & \multicolumn{2}{c}{\bfseries Copyleft} \\
\cmidrule(r){1-2}                       \cmidrule(lr){3-4}                            \cmidrule(lr){5-6}                          \cmidrule(l){7-8}
\emojibooks      & Cultural             & 49828028224 & 39.36                         &  3540432035 &     10.27                   &  641188546 &     8.08                  \\
\emojibank       & Economic             &   110278741 &  0.09                         &  \emptycell &\emptycell                   & \emptycell &\emptycell                 \\
\emojiscales     & Legal                &  2777632868 &  2.19                         &   201263562 &      0.58                   & \emptycell &\emptycell                 \\
\emojinewspaper  & News                 & 72659063648 & 57.39                         &  \emptycell &\emptycell                   &   14131835 &     0.18                  \\
\emojispeech     & Political            &   841135819 &  0.66                         &    29409301 &      0.09                   & 2684345698 &    33.84                  \\
\emojimicroscope & Scientific           &   286683148 &  0.23                         &   103499411 &      0.30                   &  444929702 &     5.61                  \\
\emojiglobe      & Web                  &   103461322 &  0.08                         & 30602455788 &     88.76                   & 4147518467 &    52.29                  \\
\bottomrule
\end{tabularx}

%% file: table-toxicity.tex
\sffamily
\begin{tabularx}{\linewidth}{@{}X*{5}{S[table-format=2.3]}@{}}
\toprule
\multirow{2}{*}{\bfseries Kind}   & \multicolumn{4}{c}{\bfseries Non-Toxic} & {\bfseries Mildly Toxic} \\
                        \cmidrule(r){2-5} \cmidrule(l){6-6}
                        & {\bfseries 0}   & {\bfseries 1}   & {\bfseries 2}   & {\bfseries 3}   &{\bfseries $\geq$ 4}\\ 
\midrule
Ability                 & 0.99  & 0.00  & 0.00  & 0.00  & 0.00      \\
Gender/Sex              & 0.98  & 0.01  & 0.01  & 0.00  & 0.00      \\
Race/Origin             & 0.94  & 0.01  & 0.04  & 0.01  & 0.00      \\
Religion                & 0.97  & 0.02  & 0.01  & 0.00  & 0.00      \\
Violence                & 0.86  & 0.11  & 0.03  & 0.01  & 0.00      \\
\bottomrule
\end{tabularx}

%% file: conf25-german-commons-part6.tex
\section{Limitations and Ethical Considerations}

The German Commons inherits fundamental limitations from its constituent sources and curation methodology. We identify four primary limitations requiring explicit acknowledgment and propose future mitigation strategies.

First, the corpus exhibits temporal bias toward historical content. News (47.02\%) and cultural domains (35.25\%) comprise 82.27\% of tokens, with cultural content predominantly sourced from 18th-20th century digitized texts. This historical skew induces nostalgia bias \cite{zhu:2024}. Scientific (0.54\%) and economic (0.07\%) domains remain critically underrepresented. Adding contemporary German text to rebalance the temporal distribution is paramount for future extensions of the corpus.
Second, the predominance of OCR-extracted text may introduce errors. German diacritics exhibit heightened vulnerability to misrecognition \cite{kanerva:2025}. While our pipeline applies OCR-specific filtering, residual errors may persist particularly in older texts. We did not apply LLM-based correction methods for error reduction, due to substantial computational expenditure, hallucination risks, and misinterpretation of historical texts, however, would be a future improvement if specialized correction models become available.

Third, standard German dominates content, diminishing linguistic diversity against non-standard varieties like Swiss, Austrian, or Low German dialects. Demographic biases may be present; socioeconomic stratification manifests through overrepresentation of formal registers from institutional sources. Cultural representation likely exhibits Western Protestant bias consistent with broader German NLP resources \cite{kurpicz:2020}. Targeted inclusion of dialectal and minority language varieties can improve this situation, if they become available under open licenses.
Finally, privacy protection through PII removal provides limited security. Our regex- and Presidio-based approaches constitute surface-level modifications. However, the historical skew and data sources from the public record diminish the potential adverse effects should PII be contained in the data.

To enable informed downstream usage, we provide comprehensive documentation following established frameworks \cite{gebru:2021,bender:2018} (\Cref{app:datasheet}). The corpus includes document-level metadata preserving provenance and license information. Publishing the deduplication bloom filters enables cross-corpus contamination detection. Thematic partitioning supports selective usage based on application requirements. Together with the corpus, we also release a highly scalable preprocessing pipeline with specific considerations for German language. This enables future additions and community contributions to the collection.

%% file: conf25-german-commons-sum.tex
\section{Conclusion}

The \emph{German Commons} is a data collection intended to address a fundamental challenge in open German language model development: the scarcity of large-scale, verifiably licensed training data for German language. By systematically aggregating \gcNumTokensInBillions~billion tokens of openly licensed German text from institutional sources, this work represents the largest collection of open German text to date and enables the development of language models without the licensing issues prevalent in web-scraped alternative corpora.

The corpus encompasses seven thematic domains, with text from web, political, legal, news, economics, cultural, and scientific sources. We apply systematic quality filtering, deduplication, and PII removal. Through detailed corpus statistics, we show the suitability of the included data for model training, and verify the high quality of the provided text. Every document in the corpus is further tagged with an explicit canonical SPDX license URL, enabling unambiguous downstream use. The \emph{German Commons} thus represent a critical step toward sustainable, ethically compliant development of German language models. 

\begin{acks}
This work has been partially funded by the German Federal Ministry of Research, Technology, and Space  (BMFTR) under Grants \textnumero~\texttt{01IS24077A}, \textnumero~\texttt{01IS24077B}, and \textnumero~\texttt{01IS24077D}; by the ScaDS.AI Center for Scalable Data Analytics and Artificial Intelligence, funded by the BMFTR and by the Sächsische Staatsministerium für Wissenschaft, Kultur und Tourismus under Grant \textnumero~\texttt{ScaDS.AI}; and by the OpenWebSearch.eu project, funded by the European Union under Grant \textnumero~\texttt{GA 101070014}.
\end{acks}

%% file: table-filter-parameters.tex
\sffamily
\begin{tabularx}{\linewidth}{@{}lXl@{}}
\toprule
\bfseries Parameter                     & \bfseries Explanation                                                                                     & \bfseries \kern-1em Excl. if \\
\midrule
Alphabetic Word Ratio                   & Ratio of whitespace-separated words consisting of only alphabetic characters                              & $> 0.99$ \\            
Bullet Line Ratio                       & Ratio of lines starting with \texttt{•} or \texttt{-} characters                                          & $> 0.70$ \\
Ellipsis Line Ratio                     & Ratio of lines ending in \texttt{...} or \texttt{…}                                                       & $> 0.30$ \\
Ellipsis Ratio                          & Ratio of \texttt{...} or \texttt{…} substrings occuring to overall whitespace-separated words             & $> 0.10$\\
Hash Ratio                              & Ratio of \texttt{\#} character occuring to overall whitespace-separated words                             & $> 0.10$\\
Stop-word Count                         & Number of stopwords in text; for stop words used, see below                                               & $< 6.00$\\  
\midrule
Duplicated Line Fraction                & Amount of duplicated lines in a document, measured as ratio of lines.                                     & $> 0.25$ \\
Duplicated Lines Character Fraction     & Amount of duplicated lines in a document, measured as ratio of characters.                                & $> 0.15$ \\
Duplicated Paragraph Fraction           & Amount of duplicated paragraphs in a document, measured as ratio of paragraphs.                           & $> 0.30$ \\
Duplicated Paragraph Character Fraction & Amount of duplicated paragraphs in a document, measured as ratio of characters.                           & $> 0.20$ \\
\midrule
Duplicate 5-gram Character Fraction     & \multirow{6}{*}{Text accounted for by duplicated n-grams, measured as ratio of characters.}               & $> 0.39$ \\
Duplicate 6-gram Character Fraction     &                                                                                                           & $> 0.39$ \\
Duplicate 7-gram Character Fraction     &                                                                                                           & $> 0.38$ \\
Duplicate 8-gram Character Fraction     &                                                                                                           & $> 0.38$ \\
Duplicate 9-gram Character Fraction     &                                                                                                           & $> 0.37$ \\
Duplicate 10-gram Character Fraction    &                                                                                                           & $> 0.37$ \\
\midrule
Top-2-Gram Character Fraction           & \multirow{3}{*}{Text accounted for by most frequent n-gram, measured as ratio of characters.}             & $> 0.07$\\
Top-3-Gram Character Fraction           &                                                                                                           & $> 0.10$\\
Top-4-Gram Character Fraction           &                                                                                                           & $> 0.13$\\
\midrule
Spacing Anomaly Ratio                  & Ratio of spacing anomalies; missing spaces, excessive spaces, spaced words                                 & $> 0.15$ \\
Case Anomaly Ratio                     & Ratio of case anomalies such as random capitalization and mixed case within words                          & $> 0.10$ \\
Word Fragment Ratio                    & Ratio of likely OCR word fragments (1-2 character words excluding common words)                            & $> 0.20$ \\
Line Artifact Ratio                    & Ratio of lines that are likely OCR artifacts (single characters, page numbers)                             & $> 0.25$ \\
Special Character Density              & Density of unusual unicode characters                                                                      & $> 0.03$ \\
Repeated Character Ratio               & Ratio of text consisting of repeated character sequences and repeated patterns                             & $> 0.05$ \\
Numeric Context Errors                 & Ratio of numbers inappropriately embedded within words (excluding ordinals)                           & $> 0.08$ \\
\midrule
Avg. Word Length (min)                 & Average length of whitespace-separated words in characters                                                 & $< 4.80$ \\  
Avg. Word Length (max)                 & Average length of whitespace-separated words in characters                                                 & $> 7.30$ \\
Word Length Standard Deviation (min)   & Standard deviation of word lengths in characters                                                           & $< 1.00$ \\
Word Length Standard Deviation (max)   & Standard deviation of word lengths in characters                                                           & $> 5.00$ \\
Very Short Words Ratio                 & Ratio of words with length $\leq$ 1 character after removing punctuation                                   & $> 0.10$ \\
Very Long Words Ratio                  & Ratio of words with length $\geq$ 15 characters after removing punctuation                                 & $> 0.10$ \\
\bottomrule
\end{tabularx}

%% file: table-stopwords.tex
\sffamily
\begin{tabularx}{\linewidth}{@{}p{5cm}X@{}}
\toprule 
\bfseries Category & \bfseries Words \\
\midrule
Definite articles           & der, die, das, den, dem, des \\
Indefinite articles         & ein, eine, einen, einem, einer \\
Conjunctions                & und, oder, aber \\
Common Verbs                & ist, sind, hat, haben, wird, werden, \\
Prepositions                & von, zu, mit, in, auf, für, bei, nach, vor, über, unter, durch, gegen, ohne, um \\  
Pronouns                    & ich, du, er, sie, es, wir, ihr, sich, sein, seine, ihrer, ihren, mich, dich \\
Adverbs                     & nicht, auch, nur, noch, schon, hier, dort, da, dann, jetzt, heute sehr, mehr, weniger, ganz, gar, etwa\\
Subordinating Conjunctions  & dass, wenn, als, wie \\
Contractions                & an, am, im, ins, zum, zur, vom, beim \\
Question words              & was, wer, wo, wann, warum, wie, welche, welcher \\
Quantifiers                 & alle, viele, einige, andere, jede, jeden, jeder \\
Modal Verbs                 & kann, könnte, muss, soll, will, würde \\
Particles                   & ja, nein, doch, so, also, nun, mal \\
\bottomrule
\end{tabularx}

%% file: table-pii-replacements.tex
\small
\sffamily
\begin{tabularx}{\linewidth}{@{}XXX@{}}
\toprule 
\bfseries PII Category  & \bfseries Generic Replacement & Comment \\ 
\midrule
Credit Card Numbers     & 4242 4242 4242 4242           & VISA testing number; valid but unused \\
IP Addresses            & 192.0.2.255                   & RFC 5737 Test Block 1                 \\
Email Addresses         & name@beispiel.de              & Example domain                        \\
Phone Numbers           & +49 123 45678910              & Invalid number with correct format    \\
IBAN Codes              & DE02 1203 0000 0000 2020 51   & DKB testing number; valid but unused  \\
URLs                    & https://www.beispiel.de       & Example domain                        \\
\bottomrule
\end{tabularx}
   

%% file: table-filtering-steps.tex
\footnotesize
\sffamily
\setlength{\tabcolsep}{2pt}
\renewcommand{\arraystretch}{1.2}
\begin{tabularx}{\linewidth}{
    @{}X
    S[
        table-format=2.3,
        tight-spacing=true,
        round-mode=places, 
        round-precision=3,
        scientific-notation=fixed, 
        fixed-exponent=6, 
        table-omit-exponent,
        table-omit-exponent,
        table-align-text-post=true,
        table-space-text-post={\,M},
    ]<{\,M}
    S[
        table-format=3.3,
        round-mode=places, 
        round-precision=3,
        scientific-notation=fixed, 
        fixed-exponent=9, 
        table-omit-exponent,
        table-align-text-post=true,
        table-space-text-post={\,B},
    ]<{\,B}
    S[
        table-format=2.3,
        tight-spacing=true,
        round-mode=places, 
        round-precision=3,
        scientific-notation=fixed, 
        fixed-exponent=6, 
        table-omit-exponent,
        table-omit-exponent,
        table-align-text-post=true,
        table-space-text-post={\,M},
    ]<{\,M}
    >{\color{gray-600}}S[
        table-format=2.2,
        tight-spacing=true,
        round-mode=places, 
        round-precision=2,
        scientific-notation=fixed, 
        fixed-exponent=0,
        table-omit-exponent,
        table-align-text-post=true,
        table-space-text-post={\,\%}, 
    ]<{\,\%}
    S[
        table-format=3.3,
        round-mode=places, 
        round-precision=3,
        scientific-notation=fixed, 
        fixed-exponent=9, 
        table-omit-exponent,
        table-align-text-post=true,
        table-space-text-post={\,B},
    ]<{{\,B}}
    >{\color{gray-600}}S[
        table-format=2.2,
        tight-spacing=true,
        round-mode=places, 
        round-precision=2,
        scientific-notation=fixed, 
        fixed-exponent=0,
        table-omit-exponent,
        table-align-text-post=true,
        table-space-text-post={\,\%}, 
    ]<{\,\%}
    S[
        table-format=2.3,
        tight-spacing=true,
        round-mode=places, 
        round-precision=3,
        scientific-notation=fixed, 
        fixed-exponent=6, 
        table-omit-exponent,
        table-omit-exponent,
        table-align-text-post=true,
        table-space-text-post={\,M},
    ]<{\,M}
    >{\color{gray-600}}S[
        table-format=2.2,
        tight-spacing=true,
        round-mode=places, 
        round-precision=2,
        scientific-notation=fixed, 
        fixed-exponent=0,
        table-omit-exponent,
        table-align-text-post=true,
        table-space-text-post={\,\%}, 
    ]<{\,\%} 
    S[
        table-format=3.3,
        tight-spacing=true,
        round-mode=places, 
        round-precision=3,
        scientific-notation=fixed, 
        fixed-exponent=9, 
        table-omit-exponent,
        table-align-text-post=true,
        table-space-text-post={\,B},
    ]<{{\,B}}
    >{\color{gray-600}}S[
        table-format=2.2,
        tight-spacing=true,
        round-mode=places, 
        round-precision=2,
        scientific-notation=fixed, 
        fixed-exponent=0,
        table-omit-exponent,
        table-align-text-post=true,
        table-space-text-post={\,\%}, 
    ]<{\,\%}
    S[
        table-format=2.3,
        tight-spacing=true,
        round-mode=places, 
        round-precision=3,
        scientific-notation=fixed, 
        fixed-exponent=6, 
        table-omit-exponent,
        table-omit-exponent,
        table-align-text-post=true,
        table-space-text-post={\,M},
    ]<{\,M}
    >{\color{gray-600}}S[
        table-format=2.2,
        round-mode=places, 
        round-precision=2,
        scientific-notation=fixed, 
        fixed-exponent=0,
        table-omit-exponent,
        table-align-text-post=true,
        table-space-text-post={\,\%}, 
    ]<{\,\%} 
    S[
        table-format=3.3,
        tight-spacing=true,
        round-mode=places, 
        round-precision=3,
        scientific-notation=fixed, 
        fixed-exponent=9, 
        table-omit-exponent,
        table-align-text-post=true,
        table-space-text-post={\,B},
    ]<{\,B}
    >{\color{gray-600}}S[
        table-format=2.2,
        tight-spacing=true,
        round-mode=places, 
        round-precision=2,
        scientific-notation=fixed, 
        fixed-exponent=0,
        table-omit-exponent,
        table-align-text-post=true,
        table-space-text-post={\,\%}, 
    ]<{\,\%} 
    @{}
}
\toprule
\multirow{2}{*}{\bfseries  Data Source}                                               
& \multicolumn{2}{c}{\bfseries Initial\textsuperscript{$\dagger$}}      
& \multicolumn{4}{c}{\bfseries Filtered}           
& \multicolumn{4}{c}{\bfseries Deduplicated}
& \multicolumn{4}{c}{\bfseries Final\textsuperscript{$\ddagger$}}      \\
\cmidrule(lr){2-3}\cmidrule(lr){4-7}\cmidrule(lr){8-11}\cmidrule(lr){12-15}
                                           & \multicolumn{1}{c}{\# Docs}
                                           & \multicolumn{1}{c}{\# Tokens}
                                           & \multicolumn{2}{c}{\# Docs}
                                           & \multicolumn{2}{c}{\# Tokens}
                                           & \multicolumn{2}{c}{\# Docs}
                                           & \multicolumn{2}{c}{\# Tokens} 
                                           & \multicolumn{2}{c}{\# Docs}
                                           & \multicolumn{2}{c}{\# Tokens}\\
\midrule
Wikipedia                                  &   2940078 &  5981495782  &   2930594 &   99.68 &  3036713451 &  50.77 & 2930224 &  99.66 &  2948736281 &  49.30 & 2930224 &  99.66 &  2948751608 &  49.30 \\
Wikivoyage                                 &     20478 &    44505758  &     20372 &   99.48 &    42027270 &  94.43 &   20372 &  99.48 &    44358611 &  99.67 &   20370 &  99.47 &    42025478 &  94.43 \\
Wiki Discussions                           &   8787192 &  2232818295  &   8523529 &   96.99 &  1231705489 &  55.16 & 8349076 &  95.01 &  1218196893 &  54.56 & 8349076 &  95.01 &  1218210917 &  54.55 \\
Youtube-Commons                            &   3311230 & 15371817395  &   2968759 &   89.66 & 14735051637 &  95.86 & 2809714 &  84.85 & 14478818029 &  94.19 & 2809714 &  84.85 & 14478850964 &  94.19 \\
One Million Posts Corpus                   &   1023853 &    98535376  &    947281 &   92.52 &    95484471 &  96.90 &  946084 &  92.40 &    94872749 &  96.28 &  946082 &  92.40 &    94872633 &  96.28 \\
The Stack                                  &  23089062 & 50004217682  &    511827 &    2.22 &  1465210992 &   2.93 &  478148 &   2.07 &  1370006500 &   2.74 &  421466 &   1.83 &  1105173228 &   2.22 \\\midrule
Reichtagsprotokolle                        &       527 &  1115271995  &       522 &   99.05 &   762561192 &  68.37 &     522 &  99.05 &   703479030 &  63.08 &     522 &  99.05 &   703495637 &  63.08 \\
German Political Speeches                  &      6685 &    29664814  &      6682 &   99.96 &    29429331 &  99.21 &    6678 &  99.90 &    29409659 &  99.14 &    6678 &  99.90 &    29409655 &  99.14 \\
C. d. Drucksachen d. dt. BT                &      5045 &   961772506  &      4841 &   95.96 &   889388322 &  92.47 &    3017 &  59.80 &   528768985 &  54.98 &    3017 &  59.80 &   528769669 &  54.98 \\
C. d. Plenarprotok. d. dt. BT              &      2755 &   560962475  &      2007 &   72.85 &   340703596 &  60.74 &    1833 &  66.53 &   316034064 &  56.34 &    1833 &  66.53 &   316034708 &  56.34 \\
EuroVoc                                    &   5440875 & 49380479671  &    424351 &    7.80 &  3136263532 &   6.35 &  245838 &   4.52 &  1987727925 &   4.03 &  245838 &   4.52 &  1988111462 &   4.03 \\\midrule
Deutsches Bundesrecht                      &      6818 &   114025703  &      3771 &   55.31 &     1055490 &   0.93 &    3217 &  47.18 &     1004259 &   0.88 &    3217 &  47.18 &     1004294 &   0.88 \\
OpenLegalData                              &    251038 &  2188155266  &    250825 &   99.91 &  2159700048 &  98.70 &  249909 &  99.55 &  1915950012 &  87.56 &  249909 &  99.55 &  1915956613 &  87.56 \\
C. d. Ents. d. BFH                         &     10885 &    80661044  &     10885 &  100.00 &    70242724 &  87.08 &   10885 & 100.00 &    67791752 &  84.05 &   10885 & 100.00 &    67791931 &  84.05 \\
Ents. d. BGH (20. Jhd.)                    &     36316 &   102538741  &     36107 &   99.42 &    98855323 &  96.41 &   36062 &  99.30 &    92841974 &  90.54 &   36062 &  99.30 &    92873390 &  90.57 \\
C. d. Ents. d. BGH                         &     77892 &   436830629  &     77842 &   99.94 &   316999669 &  72.57 &   77258 &  99.19 &   292831240 &  67.04 &   77258 &  99.19 &   292832709 &  67.04 \\
C. d. Ents. d. BVerfG                      &      8949 &    81336042  &      8943 &   99.93 &    66365848 &  81.59 &    8028 &  89.71 &    39503084 &  48.57 &    8028 &  89.71 &    39503223 &  48.57 \\
C. d. Ents. d. BpatG                       &     30866 &   256893187  &     30730 &   99.56 &   200890606 &  78.20 &   30705 &  99.48 &   185096347 &  72.05 &   30705 &  99.48 &   185099188 &  72.05 \\
C. d. Ents. d. BVerwG                      &     27200 &   182311468  &     27199 &   99.99 &   142880859 &  78.37 &   27185 &  99.94 &   123487604 &  67.73 &   27185 &  99.94 &   123487739 &  67.73 \\
C. d. amtl. E.-S. BVerfG                   &       922 &    29459559  &       919 &   99.67 &    24719629 &  83.91 &     919 &  99.67 &    24427232 &  82.92 &     919 &  99.67 &    24427294 &  82.92 \\
C. d. Ents. d. BAG                         &      5625 &    62087507  &      5624 &   99.98 &    60074120 &  96.76 &    5624 &  99.98 &    48248122 &  77.71 &    5624 &  99.98 &    48248111 &  77.71 \\
EurLEX                                     &     65000 &   205273954  &     64942 &   99.91 &   201270373 &  98.05 &   64934 &  99.90 &   201251163 &  98.04 &   64934 &  99.90 &   201263562 &  98.05 \\\midrule
Deutsches Zeitungsportal                   &  10382002 & 60140091073  &   8941729 &   86.13 & 47510671948 &  79.00 & 8076164 &  77.79 & 43865757424 &  72.94 & 8076164 &  77.79 & 43871094547 &  72.95 \\
Europeana Newspapers                       &   4125327 & 26904214707  &   3296007 &   79.90 & 20727257290 &  77.04 & 3256341 &  78.94 & 20676418130 &  76.85 & 3256341 &  78.94 & 20684418365 &  76.86 \\
Wikinews                                   &     41578 &    16237077  &     26915 &   64.73 &    14858895 &  91.51 &   23266 &  55.96 &    14222567 &  87.59 &   23266 &  55.95 &    14222520 &  87.59 \\ 
Anno                                       &   2220345 & 10004917843  &   1910430 &   86.04 &  8103983453 &  81.00 & 1910281 &  86.04 &  8103444414 &  80.99 & 1910281 &  86.04 &  8103825248 &  81.00 \\\midrule
TEDEUTenders                               &    223992 &   988617996  &     60717 &   27.11 &   197416080 &  19.97 &   57214 &  25.54 &   110651889 &  11.19 &   57214 &  25.54 &   110611112 &  11.19 \\\midrule
DiBiLit-Korpus                             &      2082 &   291048458  &      2064 &   99.14 &   224393762 &  77.10 &    2062 &  99.04 &   216391365 &  74.35 &    2062 &  99.04 &   216391448 &  74.35 \\
DiBiPhil-Korpus                            &       270 &    32903552  &       269 &   99.99 &    32529609 &  98.86 &     269 &  99.99 &    32151979 &  97.72 &     269 &  99.99 &    32151997 &  97.72 \\
Wikisource                                 &    576159 &   701282926  &    481580 &   83.58 &   573811838 &  81.82 &  240689 &  41.77 &   347765083 &  49.59 &  240689 &  41.77 &   347770430 &  49.59 \\
German-PD                                  &    131846 & 62003150877  &    124228 &   94.22 & 51126663640 &  82.46 &  123914 &  93.98 & 49322092280 &  79.55 &  123592 &  93.74 & 49333198231 &  79.34 \\
BLBooks                                    &      3958 &  2133225074  &      3799 &   95.96 &  1010926486 &  47.39 &    3715 &  93.82 &  1010903961 &  47.39 &    3714 &  93.81 &  1012047216 &  47.39 \\
SBB Fulltexts                              &   4988099 &  4367381526  &   3161745 &   63.39 &  3218092568 &  73.68 & 3127203 &  62.69 &  3181845294 &  72.85 & 3127203 &  62.69 &  3181917752 &  72.85 \\
Wikiquote                                  &     10576 &     7306180  &      8885 &   84.01 &     6940084 &  94.99 &    8612 &  81.43 &     6685547 &  91.50 &    8612 &  81.43 &     6688458 &  91.54 \\ 
MOSEL                                      &   2983769 &   387646234  &   2618447 &   87.76 &   360257215 &  92.93 & 2605569 &  87.32 &   358514248 &  92.48 & 2605569 &  87.32 &   358514283 &  92.48 \\\midrule    
Dig. d. Polytechn. Journals                &       346 &   300795201  &       346 &  100.00 &   184583664 &  61.37 &     346 & 100.00 &   180257425 &  59.93 &     346 & 100.00 &   180257799 &  59.93 \\
Wikibooks                                  &     30005 &    83052041  &     27562 &   91.86 &    56807295 &  68.40 &   27292 &  90.96 &    50433541 &  60.73 &   27292 &  90.96 &    50434996 &  60.73 \\
DOAB                                       &      2103 &   318835481  &      2051 &   97.53 &   299705353 &  94.00 &    1939 &  92.20 &   166920492 &  52.35 &    1939 &  92.20 &   166920321 &  52.35 \\
arXiv                                      &    460640 &  5746394800  &        80 &    0.02 &     1050860 &   0.02 &       8 &   0.00 &      103478 &   0.00 &       8 &   0.00 &      103478 &   0.00 \\ 
Wikiversity                                &     24995 &    34702418  &     19650 &   78.61 &    30470215 &  87.80 &   16371 &  65.50 &    27804414 &  80.12 &   16371 &  65.49 &    27802099 &  80.11 \\ 
OpenALEX                                   &    115714 &   646415745  &     49286 &   42.59 &   477649662 &  73.89 &   47733 &  41.25 &   413622361 &  64.99 &   47733 &  41.25 &   413632648 &  63.99 \\\midrule    

Total                            
&     71473097      
& 304629334058  
&     37594342 
&        52.60
& 163265663889
&        53.59 
&     35835220
&        50.14 
& 154798827407
&        50.81 
&     35778211
&        50.06 
& 154558196961
&        50.73 \\
\bottomrule 
\multicolumn{15}{@{}p{\linewidth}@{}}{Percentages indicate remaining of initial. \quad\textsuperscript{$\dagger$} After filtering using the source datasets' metadata, if available. \quad\textsuperscript{$\ddagger$} Includes PII replacement and final license filtering.}
\end{tabularx}

%% file: datasheet.tex
\section*{Datasheet: German Commons}\label{app:datasheet}

\subsection*{Motivation}

\paragraph{Why was the dataset created?}
German Commons addresses the critical gap in large-scale open German text for language model training. Existing German corpora either lack explicit licensing, contain web-scraped content of uncertain provenance, or provide insufficient scale. 

\paragraph{Has the dataset been used already?}
This represents the initial release of German Commons. No external usage has occurred prior to publication. 

\paragraph{What (other) tasks could the dataset be used for?}
Beyond language model pretraining, German Commons supports all German NLP research requiring clean, license-compliant text, multilingual model development, or linguistic analysis of German text across domains. The diverse domain coverage (legal, cultural, scientific, etc.) further enables domain-specific model development and cross-domain evaluation studies.

\paragraph{Who funded the creation of the dataset?}
Dataset compilation was supported by German and European research grants: German Federal Ministry of Research, Technology, and Space  (BMFTR) under Grants \textnumero~\texttt{01IS24077A}, \textnumero~\texttt{01IS24077B}, and \textnumero~\texttt{01IS24077D}, by the ScaDS.AI Center for Scalable Data Analytics and Artificial Intelligence, funded by the BMFTR and by the Sächsische Staatsministerium für Wissenschaft, Kultur und Tourismus under Grant \textnumero~\texttt{ScaDS.AI}, and by the OpenWeb-Search.eu project, funded by the European Union under Grant \textnumero~\texttt{GA 101070014}.  Constituent datasets originate primarily from state-funded institutions across Germany and Austria.

\subsection*{Dataset Composition}

\paragraph{What are the instances?}
Each instance represents a single German-language document with associated metadata and licensing information.

\paragraph{How many instances are there in total?}
The dataset contains 35,778,211 documents comprising 154,558,196,961 GPT-2 tokens.

\paragraph{What data does each instance consist of?}
Each instance includes: 
a unique identifier for source cross-referencing,
source dataset name, 
quality-filtered and paragraph-deduplicated raw text,
canonical SPDX license URL,
thematic domain key, 
GPT-2 token count,
perplexity score from a German Wikipedia KenLM model, 
and a OCR quality score.

\paragraph{Is there a label or target associated with each instance?}
No supervised labels exist. However, each instance contains metadata labels for thematic domain classification, licensing information, and document length statistics.

\paragraph{Is any information missing from individual instances?}
Paragraph-level deduplication may alter texts from their original form. Personally identifiable information has been systematically removed.

\paragraph{Does the dataset contain all possible instances or is it a sample (not necessarily random) of instances from a larger set?}
The dataset represents a filtered subset of source collections. Filtering removes OCR errors, extraction artifacts, and low-quality or duplicated content, creating a curated selection.

\paragraph{Are there recommended data splits?}
No predefined splits are provided. All data is intended for pretraining.

\paragraph{Are there any errors, sources of noise, or redundancies in the dataset?}
Despite quality filtering and deduplication, residual issues may remain: 
\Ni cross-corpus text duplicates from overlapping sources, and 
\Nii extraction artifacts from OCR and PDF-to-text processing.

\paragraph{Is the dataset self-contained, or does it link to or otherwise rely on external resources?} 
The dataset is self-contained and centrally downloadable. The Source dataset references provided enable reproducible reconstruction.

\subsection*{Collection Process}

\paragraph{What mechanisms or procedures were used to collect the data?}
Data collection employed multiple automated procedures: 
\Ni direct download from institutional repositories and open platforms, 
\Nii programmatic crawling via APIs where available, and 
\Niii automated text extraction from PDF and other document formats using specialized libraries. Then, the open source processing pipelines were applied for quality filtering and deduplication all sources. Validation occurred through manual inspection of sample outputs, cross-verification against source repositories, and automated consistency checks.

\paragraph{How was the data associated with each instance acquired?}
All text data represents directly observable content from original sources; no inference or derivation occurred. Metadata (licensing, thematic classification, source attribution) was extracted directly from source repository information or explicitly provided by institutional datasets. Where PDF extraction was required, raw text underwent validation against source documents to verify accuracy.

\paragraph{If the dataset is a sample from a larger set, what was the sampling strategy?}
Sampling was deterministic based on explicit criteria: 
\Ni German language content as per automated classification
\Nii explicit open licensing, 
\Niii quality thresholds, and
\Niv institutional source verification. No probabilistic sampling occurred; all content meeting inclusion criteria was retained after deduplication.

\paragraph{Who was involved in the data collection process and how were they compensated?}
Data collection was conducted by the author team using automated systems. No crowdworkers, contractors, or external annotators were employed. All processing occurred through programmatic methods without manual content creation or labeling requiring compensation.

\paragraph{Over what timeframe was the data collected? Does this timeframe match the creation timeframe of the data associated with the instances?}
Collection occurred between January and August 2025, using source dataset versions available through August 31st, 2025. The underlying content creation spans multiple centuries, representing a temporal range that significantly predates and extends beyond the collection timeframe. 

\subsection*{Data Preprocessing}

\paragraph{Was any preprocessing/cleaning/labeling of the data done?}
Comprehensive preprocessing included: 
\Ni text extraction from PDFs and OCR sources with encoding normalization,
\Nii language detection and filtering for German content, and 
\Niii quality filtering targeting digitization artifacts and extraction errors, 
\Niv paragraph-level deduplication using content hashing, 
\Nv  systematic PII removal, 
\Nvi format standardization across all source types. Thematic domain classification was applied based on source dataset.

\paragraph{Was the raw data saved in addition to the preprocessed/cleaned/labeled data?}
Raw data is not provided since all constituent source datasets remain publicly accessible through their original repositories.

\paragraph{Is the software used to preprocess/clean/label the instances available?}
All preprocessing software is open source and available at \url{https://github.com/coral-nlp/german-commons} and \url{https://github.com/coral-nlp/llmdata} , ensuring complete reproducibility of the dataset.

\paragraph{Does this dataset collection/processing procedure achieve the motivation for creating the dataset stated in the first section of this datasheet?}
Yes. The procedure successfully addresses the identified gap by: 
\Ni providing the largest collection to-date of openly licensed German text, 
\Nii enabling open German language model development without licensing uncertainties, and 
\Nii establishing reproducible methodology for future dataset construction. This directly fulfills the stated motivation of creating license-compliant, large-scale German training data.

\paragraph{How will the dataset be distributed?} 
The dataset is distributed as Parquet files through multiple public repositories for redundancy. Primary distribution occurs via Hugging Face Hub at \url{https://huggingface.co/datasets/coral-nlp/german-commons}.

\paragraph{When will the dataset be released/first distributed? What license (if any) is it distributed under?}
Public release occurred on 2025/10/14. Dataset metadata and compilation are licensed under \href{https://opendatacommons.org/licenses/by/1-0/}{ODC-BY 1.0}. Individual document texts retain their original licenses as specified in each instance's SPDX URL field, creating a heterogeneous but fully documented licensing structure.

\paragraph{Are there any copyrights on the data?}
Yes. Each document retains copyright under its original creator or institutional provider, governed by the specific license indicated in the instance metadata. The compilation itself does not claim additional copyright over constituent texts.

\paragraph{Are there any fees or access/export restrictions?}
The dataset is freely accessible without fees or registration requirements. However, users must comply with individual document licenses, which may include attribution requirements or share-alike provisions. Commercial use is permitted by all constituent licenses.

\subsection*{Dataset Maintenance}
\paragraph{Who is supporting/hosting/maintaining the dataset?}
The dataset is maintained by the authors of this report.

\paragraph{Will the dataset be updated? If so, how often and by whom?}
Updates may occur when significant new German open-source collections become available. The original authors will coordinate updates, with community contributions welcomed through the open-source pipeline.

\paragraph{How will updates be communicated?}
Updates will be announced through: 
\Ni versioned releases on hosting platforms with detailed changelogs, 
\Nii academic publication updates when substantial changes occur.

\paragraph{If the dataset becomes obsolete how will this be communicated?}
Obsolescence will be communicated through deprecation notices on all hosting platforms.

\paragraph{Is there a repository to link to any/all papers/systems that use this dataset?}
No centralized usage repository will be maintained. Usage tracking occurs through standard academic citation of the dataset paper. Users are encouraged to cite the dataset publication when reporting results or building derivative works.

\paragraph{If others want to extend/augment/build on this dataset, is there a mechanism for them to do so?}
The open-source \texttt{llmdata} pipeline enables community extensions through standardized data ingestion protocols for new sources and automated quality assessment and deduplication using established filtering criteria. Community contributions undergo review by the maintenance team.

\subsection*{Ethical Considerations}

\paragraph{Were any ethical review processes conducted?}
No formal institutional review board process was conducted. The dataset relies exclusively on pre-existing, publicly available, and explicitly licensed materials from established institutional sources. Data processing incorporated ethical considerations including systematic PII removal and exclusion of sources lacking clear licensing frameworks.

\paragraph{Does the dataset contain data that might be considered confidential?}
No. All included content derives from explicitly open-licensed institutional sources.

\paragraph{Does the dataset contain data that, if viewed directly, might be offensive, insulting, threatening, or might otherwise cause anxiety?}
Potentially yes. The dataset spans centuries of German text documents, which may include historical perspectives, political viewpoints, or language that could be considered offensive by contemporary standards. The scale and temporal range make comprehensive content moderation infeasible. Users should exercise appropriate caution.

\paragraph{Does the dataset relate to people?}
The dataset may contain publicly available information relating to individuals in various contexts including historical documents, biographical information, academic citations, and government records. 